\documentclass[pdflatex,sn-mathphys-num]{sn-jnl}

\usepackage{graphicx}%
\usepackage{svg}
\usepackage{multirow}%
\usepackage{amsmath,amssymb,amsfonts}%
\usepackage{amsthm}%
\usepackage{mathrsfs}%
\usepackage[title]{appendix}%
\usepackage{xcolor}%
\usepackage{textcomp}%
\usepackage{manyfoot}%
\usepackage{booktabs}%
\usepackage{float}%

\newcommand{\NRR}{\mathrm{NRR}}
\newcommand{\HR}{\mathrm{HR}}
\newcommand{\UPR}{\mathrm{UPR}}
\newcommand{\SMAPE}{\mathrm{SMAPE}}
\newcommand{\SMAPEcov}{\mathrm{SMAPE}^{\mathrm{cov}}}
\newcommand{\SMAPEadj}{\mathrm{SMAPE}^{\mathrm{adj}}}

\begin{document}

\title[\textsc{NutriMLLM}: Multimodal Micronutrient Analysis]{\textsc{NutriMLLM}: Multimodal Large Language Models for Dietary Micronutrient Analysis}

\author*[1]{\fnm{Runze} \sur{Yan}}\email{runze.yan@emory.edu}

\author[1]{\fnm{Minxiao} \sur{Wang}}\email{minxiao.wang@emory.edu}

\author[1]{\fnm{Jiaying} \sur{Lu}}\email{jiaying.lu@emory.edu}

\author[1]{\fnm{Darren} \sur{Liu}}\email{darren.liu@emory.edu}

\author[1]{\fnm{Xiao} \sur{Hu}}\email{xiao.hu@emory.edu}

\author[2]{\fnm{Hanqi} \sur{Luo}}\email{hanqi.luo@emory.edu}


\affil[1]{\orgdiv{Center for Data Science, Nell Hodgson Woodruff School of Nursing}, \orgname{Emory University}, \orgaddress{\street{1520 Clifton Rd NE}, \city{Atlanta}, \postcode{30322}, \state{GA}, \country{United States}}}

\affil[2]{\orgdiv{Rollins School of Public Health}, \orgname{Emory University}, \orgaddress{\street{1518 Clifton Rd NE}, \city{Atlanta}, \postcode{30322}, \state{GA}, \country{United States}}}

\abstract{Comprehensive estimation of dietary micronutrients from food images could improve clinical nutrition care, but training such models requires large multimodal datasets linking diverse foods to complete nutrient profiles. We first show that existing multimodal large language models (MLLMs), including leading proprietary models, are unreliable for this task. Across five model families and four independent evaluation benchmarks (ASA24, SNAPMe, FNDDS, and NutriBench), models frequently abstained or returned statistically implausible values. To address this gap without costly expert annotation, we repurposed a decade of population-scale 24-hour dietary recalls as structured prompts for text-to-image generation. This pipeline produced a synthetic corpus of about 1.1~million image-description-nutrient triplets, each pairing a generated food image with a complete 65-nutrient label. To our knowledge, this is the largest synthetic food-image corpus with comprehensive micronutrient annotation planned for public release upon publication. Fine-tuning Qwen3-VL (2B/4B/8B/30B) and GLM-4.6V-Flash on this corpus yielded \textsc{NutriMLLM}, the first family of vision-language models specialized for comprehensive dietary micronutrient estimation. We evaluate these models with a four-component framework that separately measures abstention, hallucination, overall usability, and per-nutrient numerical accuracy. On real food images, every \textsc{NutriMLLM} variant achieved near-complete coverage across all 65 nutrients, and the largest variant matched or exceeded proprietary baselines (GPT-5, Gemini 3, and Claude Sonnet 4.5) in accuracy on most nutrients. These results show that recall-driven synthetic supervision can make image-based comprehensive micronutrient estimation a tractable engineering problem and support dietary assessment, personalized nutrition guidance, and population-scale micronutrient surveillance.}

\keywords{dietary micronutrient assessment, multimodal large language models, food image analysis, text-to-image generation}

\maketitle

\section{Introduction}\label{sec1}

Dietary micronutrients, the vitamins and minerals required in milligram or microgram quantities, are essential for human physiology, and deficiencies in any of them carry well-characterized clinical consequences. Iron deficiency causes anemia and impairs cognitive development \cite{lozoff2006iron}; inadequate vitamin~D drives rickets and contributes to osteoporosis \cite{holick2007vitamin}; insufficient periconceptional folate increases neural-tube-defect risk \cite{mrc1991prevention}; and iodine deficiency remains the most common preventable cause of cognitive impairment globally \cite{zimmermann2008iodine}. Vitamin~B\textsubscript{12}, vitamin~A, zinc, and calcium intake exhibit similarly direct links to specific clinical outcomes \cite{stabler2013vitamin,sommer1983increased,prasad2003zinc,weaver2016calcium}. Despite this clinical importance, accurate micronutrient assessment in routine care remains intractable. Gold-standard 24-hour dietary recalls (24HRs) and food-frequency questionnaires demand trained interviewers and place a substantial cognitive burden on patients, while consumer self-tracking applications focus almost exclusively on macronutrients and calories \cite{thompson2017dietary}. The result is a persistent gap between the clinical relevance of micronutrient status and the practicality of measuring it.

\begin{figure}[!ht]
\centering
\includegraphics[width=\textwidth]{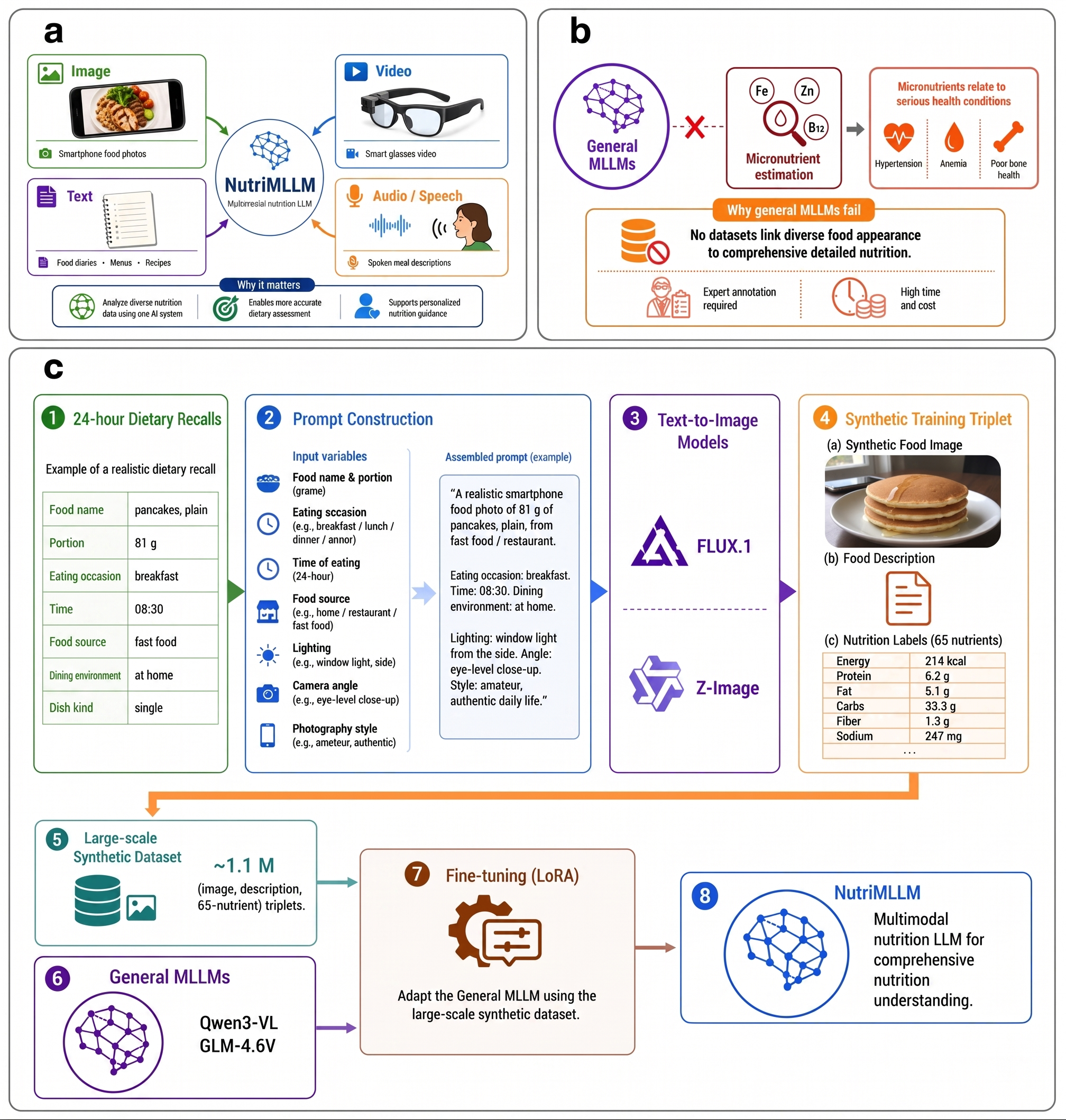}
\caption{\textbf{\textsc{NutriMLLM} motivation and training pipeline.}
\textbf{a},~Target application areas of \textsc{NutriMLLM}, integrating image, video, text, and speech inputs to support image-based dietary assessment, personalized nutrition guidance, and population-scale micronutrient surveillance.
\textbf{b},~Two limitations of general MLLMs in nutrition. (i)~Micronutrient deficiencies are linked to a wide range of clinical conditions, yet general MLLMs cannot reliably estimate micronutrient content from food images or descriptions. (ii)~Existing food datasets are either too narrow in food, meal, portion, and contextual diversity, or they lack comprehensive nutrient annotation.
\textbf{c},~Training pipeline. NHANES 24HRs supply structured food fields (name, cooking method, portion size, eating occasion, time, food source) together with FNDDS-derived 65-nutrient labels. Recall fields are combined with simulated photographic variables to drive text-to-image generation, yielding around 1.1~million image--description--nutrient triplets that LoRA-fine-tune general MLLMs into \textsc{NutriMLLM}.}\label{fig:overview}
\end{figure}

Multimodal large language models (MLLMs) are well suited to closing this gap because they do not constrain how a meal is recorded. A person can supply a photograph of a plate or a short written description, and the model works from whichever is available, rather than requiring the structured recall that makes conventional assessment burdensome. The two modalities are complementary in practice: a photograph is effortless to capture in the moment, while a description is what remains when no usable photograph was taken. A single model that handles both therefore fits the varied ways people document what they eat, which has placed MLLMs at the center of an expanding set of nutrition applications, from patient-facing dietary tracking and personalized guidance to population-scale surveillance (Fig.~\ref{fig:overview}a) \cite{lo2024dietary, khamesian2025nutrigen}. The clinically central question for any of these applications, however, is whether the model itself encodes reliable, comprehensive nutrient knowledge.

One way to sidestep this requirement is to leave the model unchanged and supply nutrient facts externally at inference, through retrieval-augmented generation (RAG) over a food-composition database \cite{yan2025dietai24}. Retrieval, however, is orthogonal to the question we study rather than an answer to it. A retrieval module can be wrapped around any backbone, a general model or a specialized one, so adding it is a separate design choice that neither establishes whether a model can itself carry comprehensive nutrient knowledge nor shows how to give it that knowledge. It also keeps the system dependent on an external database and network access at inference, working against the self-contained, on-device operation we target. We therefore treat retrieval as a complementary extension beyond the scope of this work, and ask instead whether the joint reasoning that maps a food and its portion to a complete, well-formed nutrient profile can be built into the model itself. That is what fine-tuning provides, and it is the approach we adopt.

A small but growing body of work has begun to specialize language models for nutrition through prompt engineering or supervised fine-tuning on domain corpora \cite{carrillo2025llms,gjorgjevikj2026large}. These efforts have established that MLLMs can perform food recognition, portion estimation, and macronutrient prediction from food images or text descriptions, and have introduced benchmarks for evaluating these capabilities \cite{dhaliwal2025nutribench}. The common limitation across this body of work is that supervision and evaluation cover at most four macronutrients (energy, protein, carbohydrate, and total fat), omitting the dozens of vitamins, minerals, and individual fatty acids that determine clinical micronutrient status. As a result, no published system has been trained or validated for comprehensive micronutrient prediction, and the fundamental question of whether general MLLMs encode reliable comprehensive nutrient knowledge remains both unmeasured and unaddressed. This paper closes that gap with the first systematic evaluation of MLLMs for comprehensive image-based and text-based nutrient estimation, covering five model families across four datasets. The evaluation exposes two distinct failure modes. The first, \emph{abstention}, occurs when the model is not confident enough to commit to a numeric estimate and returns no value for the nutrient in question. The second, \emph{hallucination}, occurs when the model commits to a value with apparent confidence, but the value falls outside the plausible nutritional range observed in real foods. Both failure modes are clinically consequential, though in different ways. Abstention leaves the resulting nutrient profile incomplete; because it concentrates on micronutrients, the missing entries fall precisely on the nutrients of greatest clinical interest, undermining the ability to assess deficiency status. Hallucination is the more insidious failure because the spurious value is indistinguishable from a correct one at the point of use, and can therefore propagate into clinical decisions, misclassify nutrient adequacy, or trigger inappropriate dietary and supplementation advice. Both modes are markedly worse for small open-weight MLLMs; Qwen3-VL-2B~\cite{bai2025qwen3}, for example, abstains on roughly 80\% of samples, exactly the setting most relevant to on-device dietary tracking.

The bottleneck behind these failures is annotated multimodal data. Existing public food-image datasets are typically collected from specific locations or narrow participant populations and are not nationally representative of dietary patterns, and most lack comprehensive nutrient annotation, especially for micronutrients \cite{bossard2014food, marin2021recipe1m+, thames2021nutrition5k}. Human annotation does not scale to bridge this gap: each food in a meal would require expert estimation of dozens of micronutrient values across portion sizes and preparations. National nutrition surveys, by contrast, have for decades collected 24HRs from demographically representative samples of the population. In the United States, the National Health and Nutrition Examination Survey (NHANES) links every recalled food item to a high-quality 65-nutrient profile through the Food and Nutrient Database for Dietary Studies (FNDDS) \cite{johnson2013national,bodner2006usda}, providing both the population-scale coverage and the comprehensive nutrient labeling that no public food-image dataset offers. Each recall additionally supplies the structured fields needed to describe a food (name with cooking method, portion size, eating occasion, time of day, and food source) alongside its complete nutrient label. What these surveys do not record, however, is a photograph of the food itself: the recalls supply the descriptions and the nutrient labels, but no images. We therefore use a decade of NHANES 24HRs (2013--2023) as the prompt source for text-to-image generation, augmenting each recall with simulated lighting, camera angle, and photographic style. To avoid dependence on a single generator, broaden the visual distribution of the resulting corpus, and remove the per-image cost that proprietary generators would impose at this scale, we render each prompt with two open-weight text-to-image models, Z-Image-Turbo \cite{cai2025z} and FLUX.1-dev \cite{labs2025flux1kontextflowmatching}, which carry distinct stylistic biases, yielding a synthetic multimodal corpus of around 1.1~million image–description–nutrient triplets. Fine-tuning open-weight MLLMs on this corpus, with low-rank adaptation (LoRA) \cite{hu2022lora} used for parameter and memory efficiency, transforms Qwen3-VL at 2B/4B/8B/30B \cite{bai2025qwen3} and GLM-4.6V-Flash \cite{hong2025glm} into nutrition-specialized vision-language models, which we call \textsc{NutriMLLM} (Fig.~\ref{fig:overview}). All \textsc{NutriMLLM} variants are open-weight and compact, spanning roughly 2B to 30B parameters; although the proprietary baselines do not disclose their parameter counts, frontier systems of this kind are generally believed to be at least an order of magnitude larger. This small footprint makes the variants suitable for privacy-preserving, on-device deployment, as well as for academic research and reproducible clinical evaluation. Despite this gap in scale, \textsc{NutriMLLM} achieves near-complete coverage across all 65 nutrients on the four independent evaluation datasets and matches or exceeds proprietary baselines (GPT-5 \cite{singh2025openai}, Gemini~3 \cite{google2025gemini3}, Claude Sonnet~4.5 \cite{anthropic2025sonnet45}) on abstention-adjusted accuracy on the majority of nutrients.

This work makes three contributions. First, it introduces a four-component evaluation framework that separately measures abstention, hallucination, overall usability, and numerical accuracy at the per-nutrient level, and uses this framework to deliver the first systematic validation of MLLMs for comprehensive image-based and text-based nutrient estimation. Second, it presents \textsc{NutriMLLM}, the first family of vision-language models specialized for comprehensive dietary micronutrient estimation, obtained via recall-driven synthetic supervision rather than costly expert annotation, and demonstrates that synthetic data alone are sufficient to close the nutrient-knowledge gap left by general MLLMs. Third, it releases a synthetic multimodal corpus of around 1.1~million image–description–nutrient triplets with comprehensive 65-nutrient labels, which will be released upon publication together with the training and evaluation code and fine-tuned model weights, subject to applicable model and dataset licensing constraints. Together, these contributions reposition image-based comprehensive micronutrient estimation as a tractable engineering problem and supply the model, data, and evaluation tools required to make it practical.

\section{Results}\label{sec2}

\subsection{General MLLMs fail at comprehensive nutrient estimation}\label{sec:results-validation}

We first asked whether existing MLLMs, including the leading proprietary MLLMs, already possess reliable comprehensive nutrient knowledge. The answer is the precondition for everything that follows: if proprietary MLLMs already perform reliably, no domain adaptation is needed; if they do not, the size and shape of their failure indicate where new methods (such as \textsc{NutriMLLM}) need to focus. Five MLLM families were evaluated, as described in Methods (\S\ref{sec:methods-models}): the proprietary MLLMs (GPT-5, Gemini~3, Claude Sonnet~4.5) and the general open-weight MLLMs (Qwen3-VL at 2B/4B/8B/30B and GLM-4.6V-Flash), tested on the four independent datasets described in \S\ref{sec:methods-data}. Every model received the same Prompt~2 input (Appendix~\ref{sec:appendix-prompts}) and was scored along three complementary axes corresponding to the failure modes introduced in \S\ref{sec1}. The Non-Response Rate (NRR) quantifies \emph{abstention}: the fraction of samples for which the model is not confident enough to commit to a numeric estimate and returns no value. The Hallucination Rate (HR) quantifies \emph{hallucination}: the fraction of samples for which the model commits to a value that falls outside the plausible nutritional range observed in real foods. Their sum, the Unusable Prediction Rate ($\mathrm{UPR} = \mathrm{NRR} + \mathrm{HR}$), is the overall fraction of outputs that cannot be used. The adjusted Symmetric Mean Absolute Percentage Error ($\mathrm{SMAPE}^{\mathrm{adj}}$) measures numerical accuracy with a penalty for missing predictions, correcting for the selective answering of models that abstain on harder cases. Full definitions and the rationale for these metrics are given in Appendix~\ref{sec:eval_metrics}. From a clinical-utility standpoint, a model is useful only when it provides numeric estimates across the nutrients of interest \emph{and} those estimates are accurate; abstention is therefore treated here as a meaningful failure rather than as safe behaviour, since a clinician cannot act on a missing value any more readily than on a wrong one.

\begin{figure}[H]
\centering
\includegraphics[width=\textwidth]{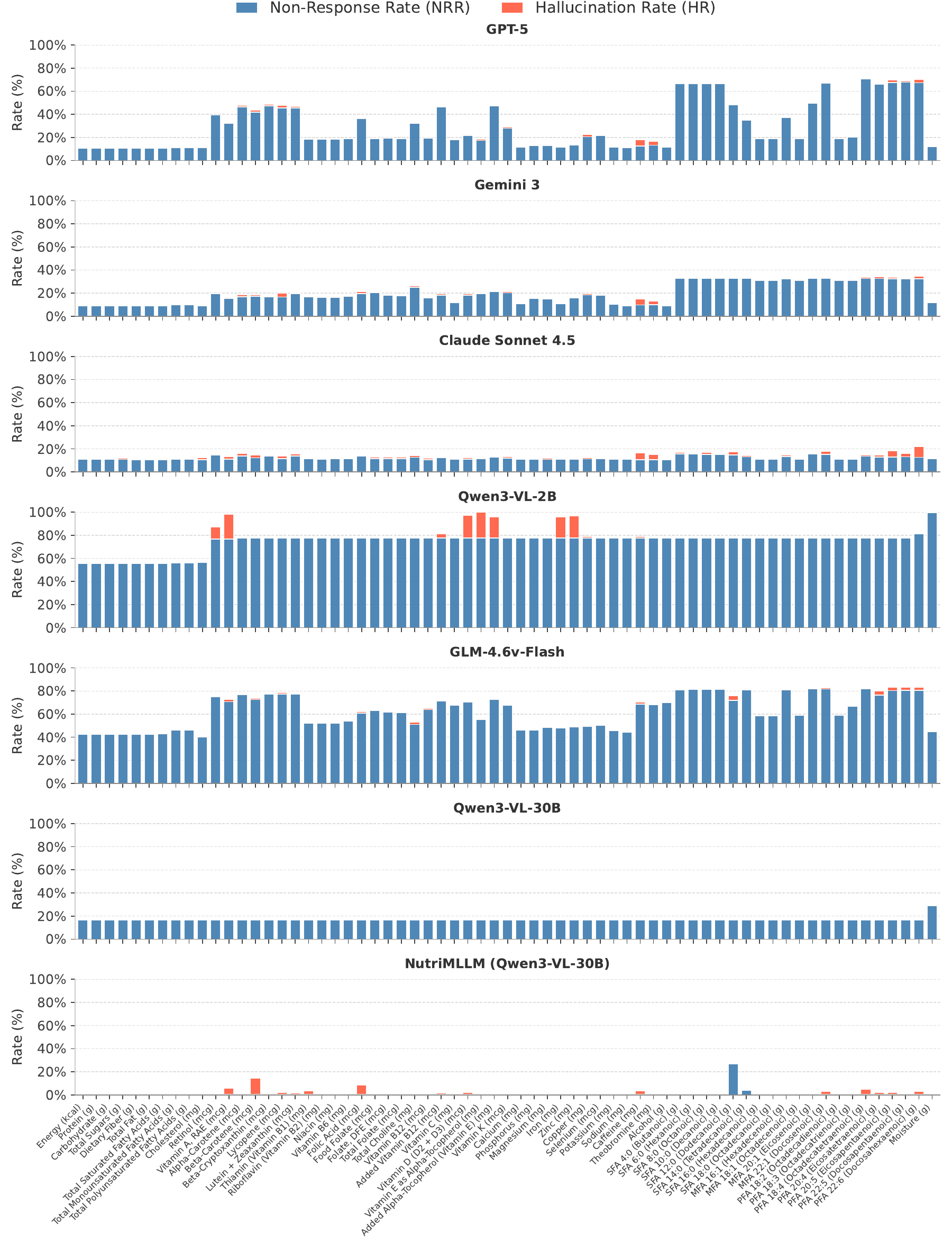}
\caption{\textbf{Non-response and hallucination rates on ASA24.}
Per-nutrient Non-Response Rate (NRR; fraction of samples for which a model abstains from producing a numeric estimate) and Hallucination Rate (HR; fraction of samples for which the predicted value falls outside the plausible nutritional range, defined as exceeding the per-nutrient 99.5th percentile of the empirical ground-truth distribution). Models are grouped into proprietary MLLMs (GPT-5, Gemini~3, Claude Sonnet~4.5), general open-weight MLLMs (Qwen3-VL at 2B/4B/8B/30B, GLM-4.6V-Flash), and \textsc{NutriMLLM} variants. General open-weight MLLMs exhibit the highest NRR, particularly on micronutrients; \textsc{NutriMLLM} reduces both NRR and HR to near zero across all 65 nutrients.}\label{fig:metric12-asa24}
\end{figure}

\begin{figure}[H]
\centering
\includegraphics[width=\textwidth]{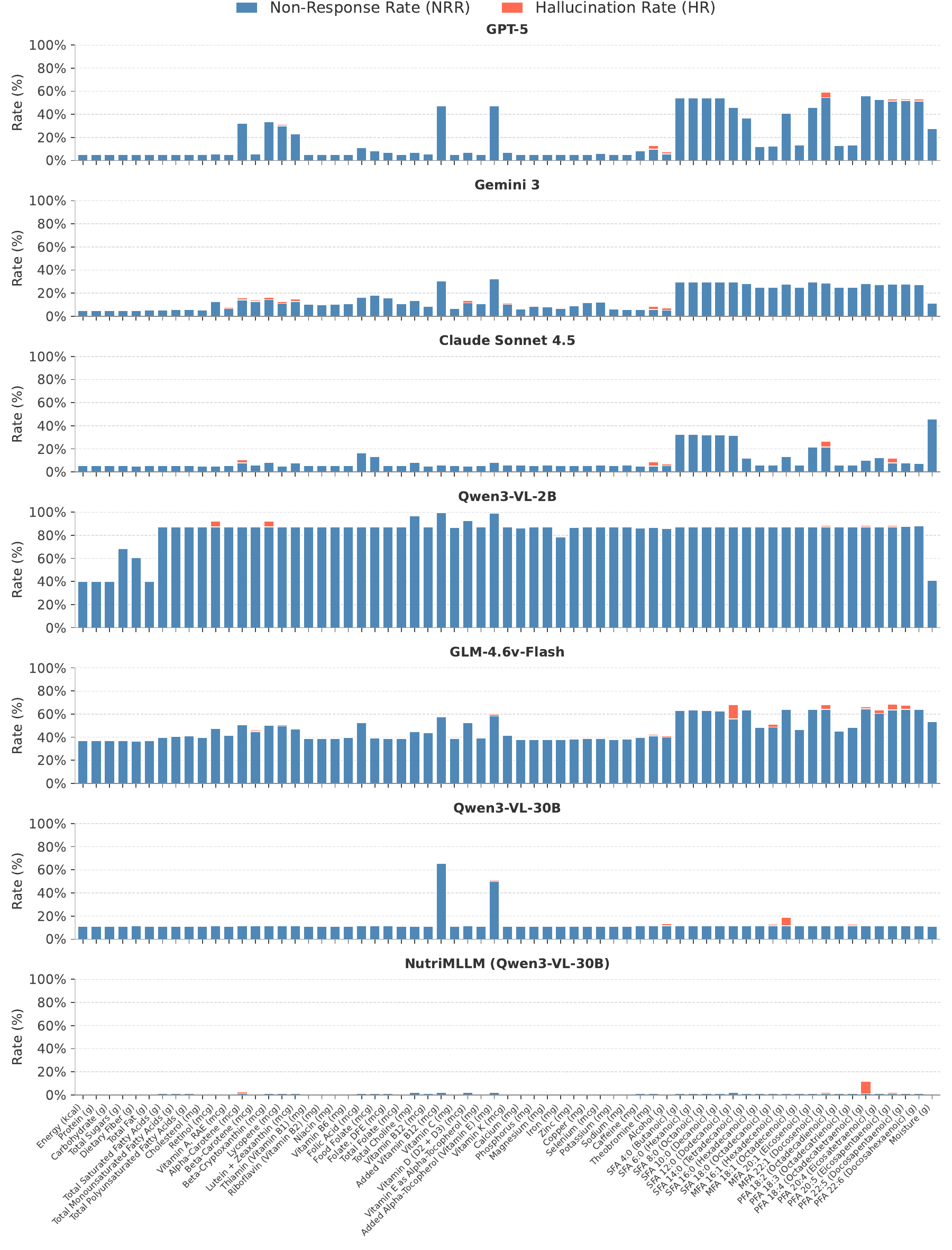}
\caption{\textbf{Non-response and hallucination rates on SNAPMe.}
Per-nutrient NRR and HR, evaluated and grouped as in Fig.~\ref{fig:metric12-asa24}. The same qualitative pattern holds on this independent set of real food photographs: general MLLMs abstain or return implausible values on a sizeable fraction of micronutrient samples, while \textsc{NutriMLLM} achieves near-complete coverage across all 65 nutrients.}\label{fig:metric12-SNAPMe}
\end{figure}

\begin{figure}[H]
\centering
\includegraphics[width=\textwidth]{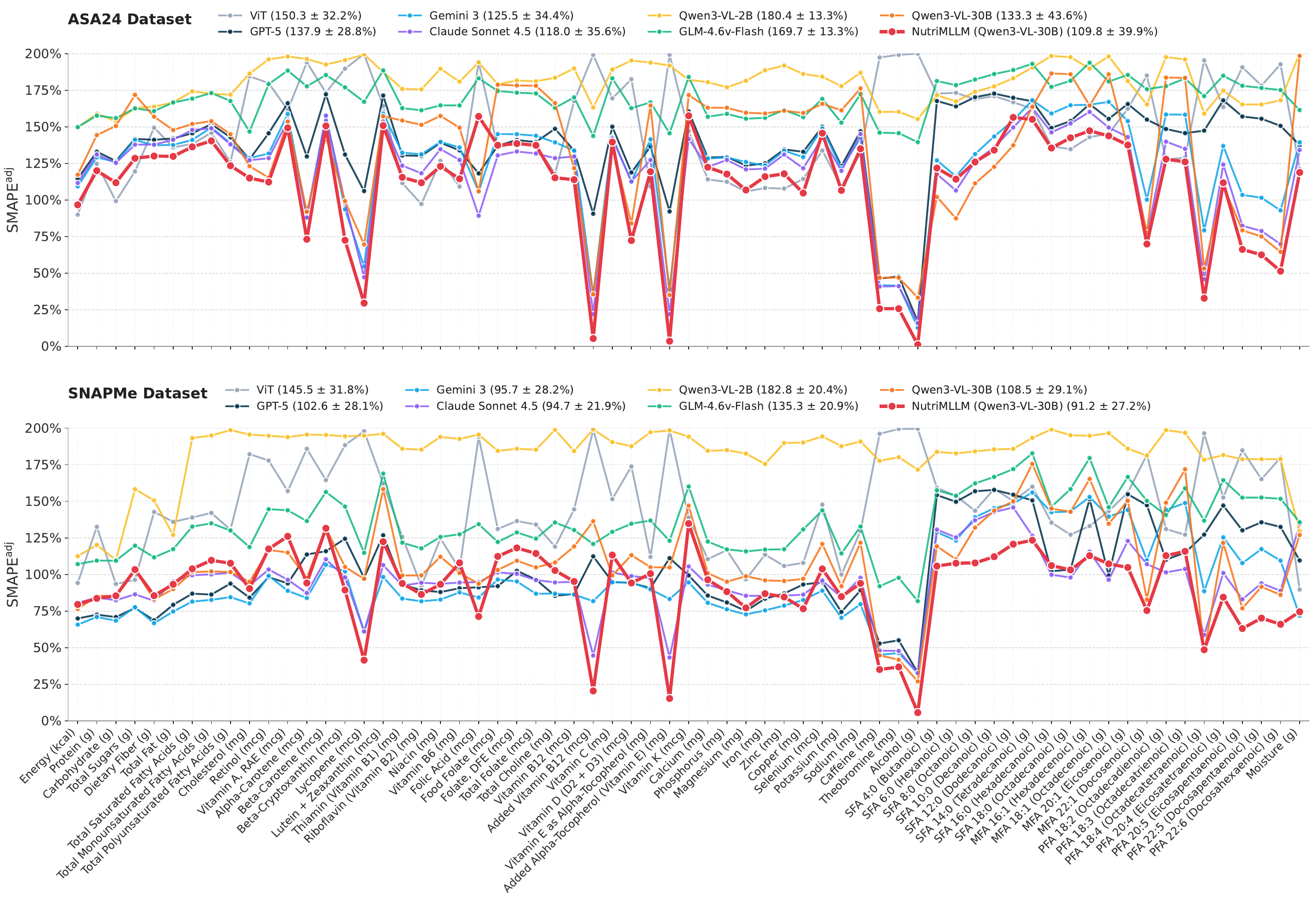}
\caption{\textbf{Per-nutrient $\mathrm{SMAPE}^{\mathrm{adj}}$ on ASA24 and SNAPMe.}
$\mathrm{SMAPE}^{\mathrm{adj}}$ measures numerical accuracy with a penalty for missing predictions, correcting for selective answering (definition and rationale in Appendix~\ref{sec:eval_metrics}). Lower is better. \textsc{NutriMLLM}-Qwen3-VL-30B matches or improves on the proprietary MLLMs on $\mathrm{SMAPE}^{\mathrm{adj}}$ for the majority of nutrients, demonstrating that synthetic supervision transfers to real food images.}\label{fig:metric3-asa24-SNAPMe}
\end{figure}

Per-nutrient NRR and HR on ASA24 and SNAPMe are reported in Figs.~\ref{fig:metric12-asa24} and \ref{fig:metric12-SNAPMe}, with each panel grouping the models into proprietary MLLMs, general open-weight MLLMs, and \textsc{NutriMLLM} variants so that within-group and between-group differences can be read off side by side; the \textsc{NutriMLLM} rows are introduced in \S\ref{sec:results-nutrimllm}, and the present subsection focuses on the general-MLLM rows. Figure~\ref{fig:metric3-asa24-SNAPMe} reports per-nutrient $\mathrm{SMAPE}^{\mathrm{adj}}$ on the same two datasets. Per-nutrient analyses for the two text-only datasets are reported in full in Appendix~\ref{sec:appendix-text-results} (Figs.~\ref{fig:metric12-fndds},~\ref{fig:metric3-fndds}, Table~\ref{tab:nutrimllm-results}); the key text-side findings are summarized below.

On both image datasets, every general MLLM, including the leading proprietary MLLMs, produced unusable predictions for a substantial fraction of the 65 nutrients. On ASA24 (Fig.~\ref{fig:metric12-asa24}), the three proprietary MLLMs exhibited markedly different failure profiles, indicating that different commercial systems carry different nutrient-knowledge structures. Claude Sonnet~4.5 stayed below 15\% NRR on nearly every nutrient but produced intermittent hallucinations on individual micronutrients and on the longest-chain polyunsaturated species. Gemini~3 and GPT-5, by contrast, concentrated their failures on the same set of nutrients, namely the individual saturated, monounsaturated, and polyunsaturated fatty-acid species (SFA, MFA, and PFA breakdowns; for example, SFA 4:0/6:0/8:0, MFA 20:1/22:1, and the long-chain PFAs 20:4, 20:5, 22:5, and 22:6), but at very different magnitudes: Gemini~3 reached 30--35\% NRR on these species, whereas GPT-5 reached 60--70\%. The general open-weight MLLMs were worse by a large margin: Qwen3-VL-2B abstained on more than 55\% of samples on every macronutrient and on 75--100\% of samples on most micronutrients; GLM-4.6V-Flash showed about 40\% NRR on macronutrients and 50--80\% on micronutrients; only Qwen3-VL-30B kept NRR below 20\% across most nutrients. HRs were lower than abstention rates overall but showed the same micronutrient bias: when a general open-weight MLLM did emit a value for a micronutrient, that value disproportionately fell into the implausible range. SNAPMe (Fig.~\ref{fig:metric12-SNAPMe}) replicated these patterns on a wholly independent set of real-life food photographs, ruling out an ASA24-specific artefact.

Within this overall failure, the macronutrient/micronutrient asymmetry was pronounced. For every general MLLM, both NRR and HR were substantially lower on the four macronutrients (energy, protein, carbohydrate, total fat) than on vitamins, minerals, and individual fatty-acid breakdowns; the proprietary MLLMs in particular kept NRR below 10\% on macronutrients but rose up to 70\% on selected micronutrients. This asymmetry is consistent with the relative abundance of macronutrient information on the public web compared with the structured micronutrient profiles required for fine-grained estimation, and it directly motivates the construction of a large-scale multimodal training corpus that carries comprehensive micronutrient labels by design.

At every metric, the proprietary MLLMs outperformed the general open-weight MLLMs. Mean $\mathrm{SMAPE}^{\mathrm{adj}}$ across all 65 nutrients on ASA24 ranged from 118.0\% (Claude Sonnet~4.5) to 137.9\% (GPT-5) for the proprietary MLLMs, compared with 133.3\% (Qwen3-VL-30B, the strongest general open-weight MLLM), 169.7\% (GLM-4.6V-Flash), and 180.4\% (Qwen3-VL-2B); on SNAPMe, the corresponding values were 94.7\%--102.6\% for the proprietary MLLMs and 108.5\%, 135.3\%, and 182.8\% for the same three general open-weight MLLMs (Fig.~\ref{fig:metric3-asa24-SNAPMe}). The ordering reflects the proprietary MLLMs' larger parameter counts and broader pre-training corpora, but the absolute values show that no general MLLM produces reliably accurate estimates across the 65-nutrient panel: even the strongest general open-weight MLLM, Qwen3-VL-30B, had mean $\mathrm{SMAPE}^{\mathrm{adj}}$ above 100\% on both datasets, and the smaller open-weight MLLMs combined high abstention with high error, the worst combination from a clinical-utility standpoint.

The same failure pattern recurred when models were given food-text descriptions instead of images, but with substantially lower magnitudes. On FNDDS (Fig.~\ref{fig:metric3-fndds}), which queries each model on a single food name across the full 65-nutrient panel, mean $\mathrm{SMAPE}^{\mathrm{adj}}$ for the proprietary MLLMs ranged from 49.2\% (Gemini~3) to 75.2\% (GPT-5), markedly better than their 118.0\%--137.9\% range on ASA24 images, and the general open-weight MLLMs improved correspondingly (Qwen3-VL-30B: 102.8\% on FNDDS vs 133.3\% on ASA24; Qwen3-VL-2B: 121.5\% vs 180.4\%; GLM-4.6V-Flash: 141.8\% vs 169.7\%). On NutriBench (Table~\ref{tab:nutrimllm-results}), even the easier four-macronutrient setting still exposed a substantial gap in the general open-weight MLLMs: the strongest general open-weight MLLM tested on this benchmark, Qwen3-VL-8B, achieved UPR of 5.9--6.7\% across the four macronutrients, but GLM-4.6V-Flash sat near 25\% and Qwen3-VL-2B exhibited UPR of 37.5--39.2\%, whereas the proprietary MLLMs remained at 0.1--0.8\%. Two conclusions follow. First, that text inputs are uniformly easier than images for every model shows that visual food recognition and portion estimation contribute their own errors to the image-based task. Second, that non-trivial UPR and $\mathrm{SMAPE}^{\mathrm{adj}}$ persist even on text-only inputs, where no visual perception is required, shows that the underlying deficit is fundamentally one of \emph{nutrient knowledge}. The clinical implication is direct: deploying current MLLMs, and especially their small edge-deployable variants, for image- or text-based comprehensive micronutrient assessment without further adaptation is not yet reliable. This is the gap that \textsc{NutriMLLM} is designed to close.

\subsection{\textsc{NutriMLLM} closes the nutrient-knowledge gap}\label{sec:results-nutrimllm}

Having established that existing general MLLMs do not yet possess reliable nutrient knowledge, the next question is whether NHANES-driven synthetic supervision can repair this deficit without any further human annotation. The aim is twofold: (i)~to demonstrate that the recall-driven training recipe described in Methods (\S\ref{sec:methods-synth},~\S\ref{sec:methods-finetune}) actually produces specialized models capable of comprehensive nutrient estimation, and (ii)~to quantify how much of the gap to the proprietary MLLMs these specialized models can close, including for the small open-weight backbones most relevant to on-device deployment. One \textsc{NutriMLLM} variant was trained for each open-weight backbone (Qwen3-VL-2B/4B/8B/30B and GLM-4.6V-Flash) on the same synthetic corpus, then re-evaluated alongside the general open-weight MLLMs and the proprietary MLLMs.

Because every general-MLLM evaluation in \S\ref{sec:results-validation} used the same figures and table, the corresponding \textsc{NutriMLLM} curves and rows are visible directly within Figs.~\ref{fig:metric12-asa24}, \ref{fig:metric12-SNAPMe}, \ref{fig:metric3-asa24-SNAPMe}, \ref{fig:metric12-fndds}, \ref{fig:metric3-fndds}, and Table~\ref{tab:nutrimllm-results}. Pre- and post-fine-tune curves can therefore be read directly off a single panel for each backbone, making the per-backbone gain immediately legible.

\begin{figure}[H]
\centering
\includegraphics[width=\textwidth]{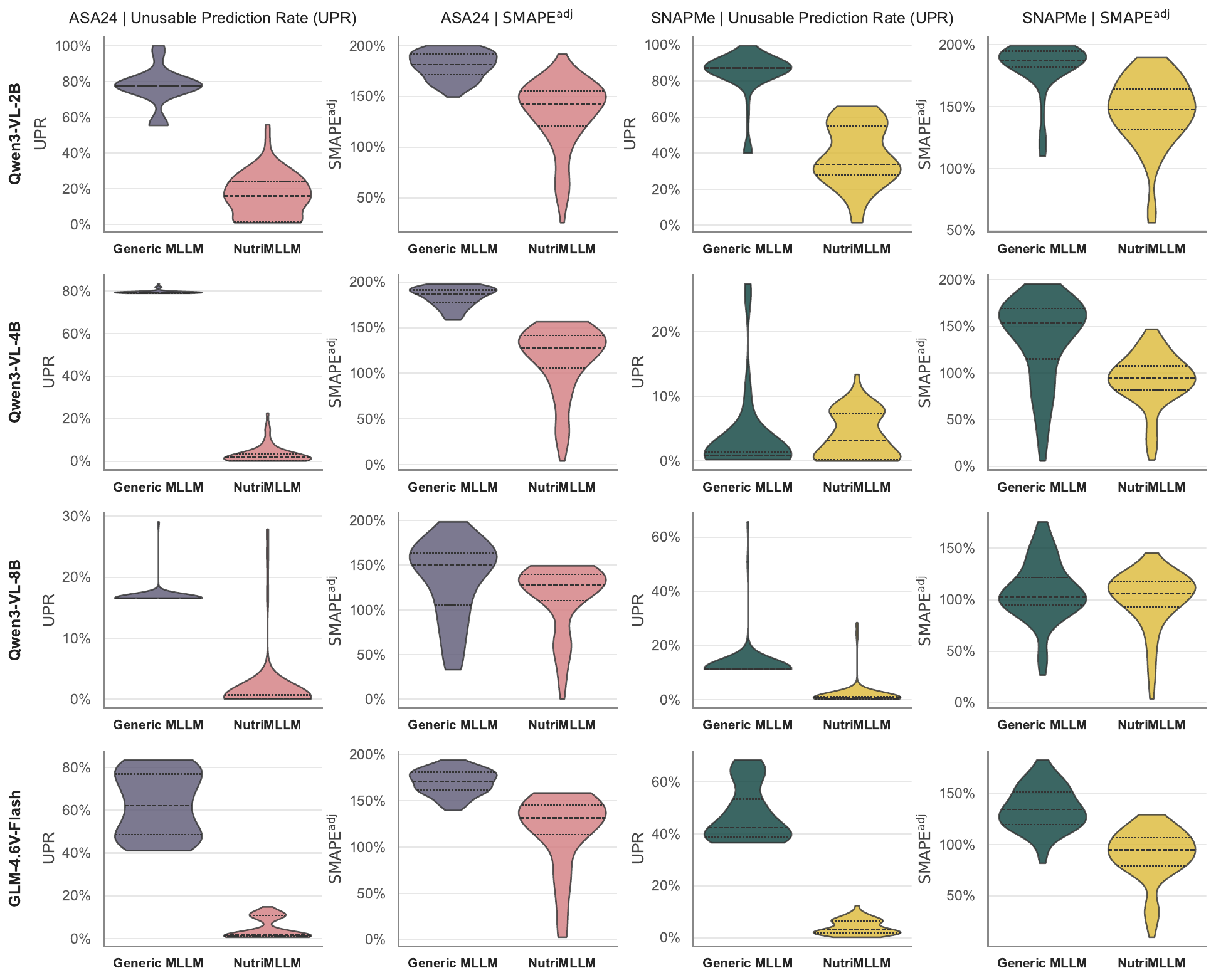}
\caption{\textbf{Per-nutrient distributions of UPR and $\mathrm{SMAPE}^{\mathrm{adj}}$ before and after LoRA fine-tuning.}
For each backbone, paired violins show the distribution across the 65 nutrients of Unusable Prediction Rate (UPR; left columns) and $\mathrm{SMAPE}^{\mathrm{adj}}$ (right columns), comparing the general (pre-fine-tune) model with the corresponding \textsc{NutriMLLM} variant on the independent evaluation set. Each violin summarizes 65 per-nutrient values. Fine-tuning shifts both distributions toward lower values across the majority of nutrients, with the largest mass shifts on micronutrients, consistent with micronutrient knowledge being the largest pre-existing deficit in general MLLMs.}\label{fig:before-after-violin}
\end{figure}

\paragraph{Coverage gain on real food images.} On both image datasets, every \textsc{NutriMLLM} variant collapsed both NRR and HR across the 65-nutrient panel, including the micronutrients on which general MLLMs had failed most severely. Per-backbone distributions in Fig.~\ref{fig:before-after-violin} show that on ASA24, median UPR across the 65 nutrients dropped from $\sim$78\% to $\sim$15\% for the 2B backbone, from $\sim$80\% to $\sim$5\% for the 4B, from $\sim$17\% to near 0\% for the 8B, and from $\sim$60\% to $\sim$3\% for GLM-4.6V-Flash; the same monotonic shifts were observed on SNAPMe. The 2B variant retained the highest residual UPR among the \textsc{NutriMLLM} models (median $\sim$15\% on ASA24, $\sim$37\% on SNAPMe), reflecting the capacity limit of the smallest backbone; the 4B, 8B, and GLM-4.6V-Flash variants all reached near-zero median UPR after fine-tuning. For the largest variant, \textsc{NutriMLLM} (Qwen3-VL-30B), NRR and HR were near zero on the bulk of the 65-nutrient panel (Figs.~\ref{fig:metric12-asa24},~\ref{fig:metric12-SNAPMe}), with a single residual NRR spike at the saturated fatty acid SFA 12:0 ($\sim$25\%) on ASA24 and one HR spike at the polyunsaturated fatty acid PFA 18:4 ($\sim$12\%) on SNAPMe. The absolute coverage of \textsc{NutriMLLM} (Qwen3-VL-30B) nonetheless remained the highest of any model evaluated, surpassing all three proprietary MLLMs across the 65-nutrient panel on both datasets.

\paragraph{Accuracy gain on real food images.} The broad coverage gain did not come at the cost of accuracy. The mean $\mathrm{SMAPE}^{\mathrm{adj}}$ of \textsc{NutriMLLM} (Qwen3-VL-30B) on ASA24 fell from 133.3\% (general Qwen3-VL-30B) to 109.8\%, below all three proprietary MLLMs (Claude Sonnet~4.5: 118.0\%; Gemini~3: 125.5\%; GPT-5: 137.9\%); on SNAPMe, it fell from 108.5\% to 91.2\%, again below the same three proprietary MLLMs (94.7\%, 95.7\%, and 102.6\% respectively) (Fig.~\ref{fig:metric3-asa24-SNAPMe}). The per-backbone violin distributions in Fig.~\ref{fig:before-after-violin} confirm the same pattern at smaller scales: $\mathrm{SMAPE}^{\mathrm{adj}}$ distributions shifted to lower values on ASA24 for all four smaller backbones, and on SNAPMe for three of the four (the 8B's SNAPMe $\mathrm{SMAPE}^{\mathrm{adj}}$ remained approximately flat while its UPR dropped substantially). \textsc{NutriMLLM} (Qwen3-VL-30B) also outperformed the Vision Transformer (ViT) baseline \cite{dosovitskiy2020image}, a traditional vision-transformer-plus-regression-head approach fine-tuned on the same NHANES-derived synthetic supervision, by a wide margin on both datasets (ViT: 150.3\% on ASA24, 145.5\% on SNAPMe). This comparison is informative because ViT and \textsc{NutriMLLM} had identical training signals, so the gap reflects the architectural advantage of the multimodal language model over a traditional regression pipeline. Per-nutrient, the $\mathrm{SMAPE}^{\mathrm{adj}}$ of \textsc{NutriMLLM} (Qwen3-VL-30B) fell below that of every proprietary MLLM on the majority of the 65 nutrients on both datasets, with the largest accuracy gains concentrated on the micronutrients where the original deficit was largest.

\paragraph{Generalization to text benchmarks.} The same qualitative pattern recurred on the food-text datasets. On NutriBench (Table~\ref{tab:nutrimllm-results}), \textsc{NutriMLLM} (Qwen3-VL-30B) achieved 0.0\% UPR on energy, the lowest of any model in the table, and the lowest UPR among non-proprietary models on protein (0.4\%), carbohydrate (0.4\%), and total fat (0.3\%). Its $\mathrm{SMAPE}^{\mathrm{adj}}$ was the best among non-proprietary models on three of the four nutrients (energy: 47.4\%; carbohydrate: 46.3\%; total fat: 54.8\%); on protein it reached 48.9\%, within a few percentage points of the best non-proprietary value of 43.8\% achieved by \textsc{NutriMLLM} (Qwen3-VL-8B). BERT \cite{devlin2019bert} in Table~\ref{tab:nutrimllm-results} is the text-side
counterpart of the ViT baseline above: a traditional
supervised model, a BERT encoder with a
regression head, fine-tuned on the same NHANES-derived synthetic
corpus as \textsc{NutriMLLM}. Because BERT is a regression model rather than a generation model, it always produces a numeric output, so its non-zero UPR (0.1--4.5\% across the four nutrients) reflects hallucination rather than abstention. Despite both architectures having access to identical training supervision, \textsc{NutriMLLM} (Qwen3-VL-30B) achieved lower UPR on every nutrient and lower $\mathrm{SMAPE}^{\mathrm{adj}}$ on every nutrient than BERT (47.4 vs 56.6 for energy, 48.9 vs 77.5 for protein, 46.3 vs 65.0 for carbohydrate, 54.8 vs 76.1 for total fat). The cleanest head-to-head against BERT is on UPR, where \textsc{NutriMLLM} (Qwen3-VL-30B) leads on all four nutrients; the $\mathrm{SMAPE}^{\mathrm{adj}}$ gap is larger but partly reflects BERT's residual UPR, which inflates its per-nutrient penalty. The smaller \textsc{NutriMLLM} variants showed the same qualitative UPR gain: \textsc{NutriMLLM} (Qwen3-VL-2B) reduced UPR from 37.5--39.2\% (general Qwen3-VL-2B) to 2.6--4.4\% across the four macronutrients, restoring practical usability to the model size targeted at edge deployment. On FNDDS (Figs.~\ref{fig:metric12-fndds},~\ref{fig:metric3-fndds}), mean $\mathrm{SMAPE}^{\mathrm{adj}}$ across the 65 nutrients fell from 102.8\% (general Qwen3-VL-30B) to 59.4\% for \textsc{NutriMLLM} (Qwen3-VL-30B), approaching the proprietary MLLMs (Gemini~3: 49.2\%; Claude Sonnet~4.5: 54.8\%) and substantially below both GPT-5 (75.2\%) and the BERT baseline (130.4\%).

\paragraph{Scaling across backbones.} Within the \textsc{NutriMLLM} family, performance scaled monotonically with backbone size on both axes. On NutriBench (Table~\ref{tab:nutrimllm-results}), mean UPR across the four macronutrients dropped from 3.2\% at the 2B scale to 1.5\% at 4B, 0.6\% at 8B, and 0.3\% at 30B; mean $\mathrm{SMAPE}^{\mathrm{adj}}$ dropped from 102\% at 2B to 68\% at 4B, 55\% at 8B, and 49\% at 30B. The same monotonic ordering held on the image datasets, with Fig.~\ref{fig:before-after-violin} showing progressively lower \textsc{NutriMLLM} UPR and $\mathrm{SMAPE}^{\mathrm{adj}}$ distributions for the 2B/4B/8B/GLM-4.6V-Flash backbones evaluated there, and Fig.~\ref{fig:metric3-asa24-SNAPMe} placing the 30B at the lowest mean $\mathrm{SMAPE}^{\mathrm{adj}}$ of any model. The two ends of the ladder therefore serve complementary deployment regimes: the small variants are most useful where on-device inference is required for privacy or connectivity reasons (\textsc{NutriMLLM} (Qwen3-VL-2B) remains the variant on which fine-tuning most dramatically changes practical viability, reducing UPR from roughly 39\% to 3\% on NutriBench and from a median of roughly 78\% to 15\% on ASA24), while the large variant is most useful where institutional compute is available and frontier-level accuracy is required (\textsc{NutriMLLM} (Qwen3-VL-30B) exceeds every proprietary MLLM on $\mathrm{SMAPE}^{\mathrm{adj}}$ on the majority of nutrients across both image and text benchmarks).

In summary, recall-driven synthetic supervision is sufficient to convert general open-weight MLLMs into nutrition specialists. On real food images, \textsc{NutriMLLM} (Qwen3-VL-30B) achieved the best mean $\mathrm{SMAPE}^{\mathrm{adj}}$ of any model evaluated (109.8\% on ASA24, 91.2\% on SNAPMe), ahead of all three proprietary MLLMs and of the ViT traditional ML baseline (150.3\% on ASA24, 145.5\% on SNAPMe) fine-tuned on the same synthetic supervision. On text benchmarks, \textsc{NutriMLLM} (Qwen3-VL-30B) approaches the proprietary MLLMs on FNDDS (59.4\% vs Gemini~3's 49.2\%) and surpasses the BERT traditional ML baseline (also fine-tuned on the same synthetic supervision) on every metric on NutriBench. The pattern holds across model sizes (2B to 30B parameters), modalities (image and text), and dataset granularities (single foods and full meals). The gain is consistently largest exactly where the original deficit was largest, namely on micronutrients and on small open-weight backbones, which are the regimes that matter most for clinical and on-device applications.

\subsection{Ablations and design choices}\label{sec:results-ablations}

Sections~\ref{sec:results-validation} and \ref{sec:results-nutrimllm} establish that recall-driven synthetic supervision converts general open-weight MLLMs into reliable nutrient estimators. This section examines why the recipe works and validates the design choices behind it. Three questions are addressed in turn: whether the coverage gain reflects genuine knowledge acquisition rather than uninformed guessing; whether training on images from multiple text-to-image generators improves downstream performance; and how validation performance evolves over training, which determines the checkpoint used for all evaluations reported in the paper.

\paragraph{The coverage gain reflects genuine knowledge acquisition.} The coverage gains of \S\ref{sec:results-nutrimllm} could in principle be hollow: because a model that emits a value for every nutrient regardless of accuracy would also show near-zero UPR, the improvement might reflect a shift from abstention to uninformed guessing rather than genuine knowledge. Figure~\ref{fig:before-after-violin} shows that it does not. Indiscriminate guessing would push the post-fine-tune $\mathrm{SMAPE}^{\mathrm{adj}}$ distribution toward larger errors, yet across all four backbones the distribution instead shifts toward lower values or remains stable, never developing an inflated upper tail. The shift is also nutrient-specific in exactly the way genuine knowledge transfer predicts: the largest accuracy gains fall on the micronutrients where general MLLMs had failed most severely, while macronutrients, already handled well, change little. Guessing would degrade the hardest newly-answered nutrients rather than improve them, and would not concentrate its effect on the previously-deficient nutrients. The coincidence of the largest gains with the largest pre-existing deficits is therefore the signature of real, previously-absent nutrient knowledge being supplied by the synthetic supervision.

\begin{figure}[H]
\centering
\includegraphics[width=\textwidth]{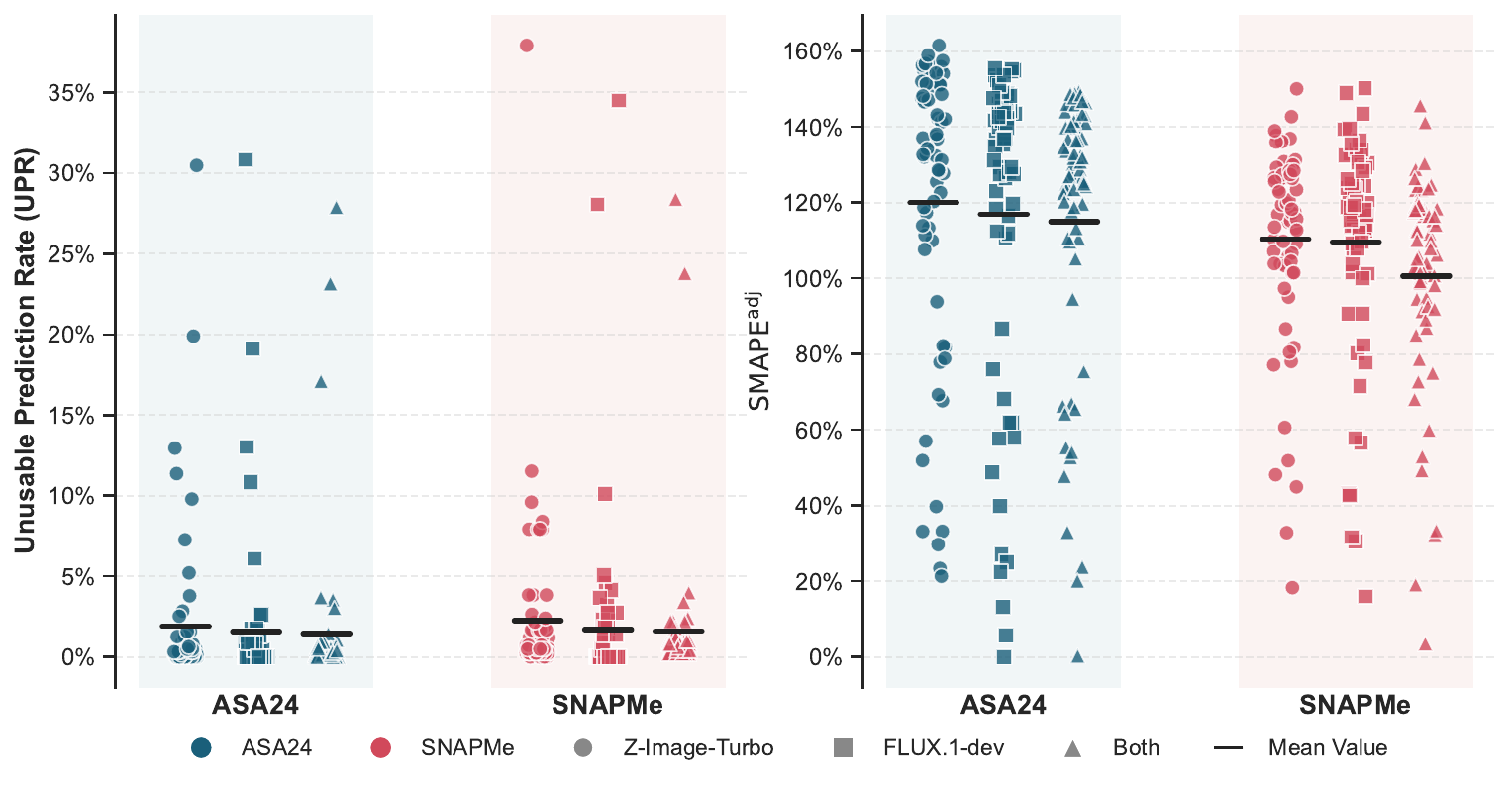}
\caption{\textbf{Effect of the synthetic-image generator on downstream \textsc{NutriMLLM} performance.}
Each marker is a \textsc{NutriMLLM} variant trained on synthetic images from Z-Image-Turbo only, FLUX.1-dev only, or the union of both; horizontal lines mark per-group means. The variant trained on the union achieves the lowest UPR and $\mathrm{SMAPE}^{\mathrm{adj}}$ on both ASA24 and SNAPMe, with the largest gain on the independent SNAPMe photographs. Combining generators therefore improves both accuracy and generalization to real food images.}\label{fig:diff-image-generators}
\end{figure}

\paragraph{Combining multiple generators improves performance and generalization.} A central design choice is the use of synthetic images from more than one text-to-image generator. Figure~\ref{fig:diff-image-generators} compares \textsc{NutriMLLM} variants trained on images from Z-Image-Turbo only, FLUX.1-dev only, and the union of both. The variant trained on the union achieved the lowest UPR and the lowest $\mathrm{SMAPE}^{\mathrm{adj}}$ of the three on both evaluation datasets. The advantage was modest on ASA24 (mean $\mathrm{SMAPE}^{\mathrm{adj}}$ of around 115\% for the union versus 117--120\% for the single-generator variants) but larger on SNAPMe, where mean $\mathrm{SMAPE}^{\mathrm{adj}}$ fell to around 100\% for the union against 109--110\% for either generator alone. Each generator imposes a distinct visual style on the synthetic food images (Fig.~\ref{fig:generated-images}), so training on the union exposes the model to a broader range of food appearances; that the larger gain appears on the independent SNAPMe photographs indicates this diversity improves generalization to real images, not merely in-sample accuracy. Generator diversity is therefore a genuine contributor to the recipe rather than an incidental choice.

\paragraph{Validation performance improves over the first epochs of training.} Figure~\ref{fig:checkpoint-comparison} tracks validation UPR and $\mathrm{SMAPE}^{\mathrm{adj}}$ on the two independent datasets across five training epochs over the synthetic corpus. UPR is low from the first epoch and stays low throughout, while $\mathrm{SMAPE}^{\mathrm{adj}}$ improves over the early epochs: median $\mathrm{SMAPE}^{\mathrm{adj}}$ falls from around 135\% to 128\% on ASA24 and from around 111\% to 106\% on SNAPMe between epochs~1 and~3. Checkpoint~3, which offers the best balance of UPR and $\mathrm{SMAPE}^{\mathrm{adj}}$ across the two independent datasets. The steady early-epoch improvement indicates that the NHANES-derived synthetic supervision is dense and well-aligned with the downstream nutrient-estimation task, a property that would not hold under noisy or weak supervision. From a practical standpoint, three epochs over this corpus is computationally modest: a research group with a single GPU node can reproduce \textsc{NutriMLLM} in around one to two days of wall-clock training.

Together, these analyses show that \textsc{NutriMLLM}'s gains reflect genuine nutrient-knowledge acquisition rather than a shift from abstention to guessing, that training on synthetic images from multiple generators improves both accuracy and generalization to real food photographs, and that strong validation performance is reached within a few epochs of modest single-node training. The recall-driven synthetic-supervision recipe is therefore both effective and practically reproducible.

\section{Methods}\label{sec:methods}

\subsection{Evaluation datasets}\label{sec:methods-data}

\textsc{NutriMLLM} and all baseline models were evaluated on four independent datasets, none of which contributed any example to training. The four were chosen to span the two modalities a nutrition MLLM must handle and, within each, to separate the sub-capabilities that comprehensive nutrient estimation depends on. ASA24 and SNAPMe are image datasets; NutriBench and FNDDS are text datasets. Together they cover controlled and in-the-wild capture, single foods and full meals, and macronutrient-only and comprehensive 65-nutrient supervision, so that performance can be attributed to a specific capability rather than to a quirk of one benchmark. Ground truth for ASA24, SNAPMe, and FNDDS is the full 65-nutrient FNDDS profile; NutriBench provides four macronutrient labels.

\paragraph{ASA24.} The Automated Self-Administered 24-Hour Dietary Assessment Tool (ASA24) Portion Size Image Database, released by the US National Cancer Institute, contains over 17,000 food images, each linked to a specific FNDDS food portion \cite{asa24portionimages}. Every image therefore carries an exact 65-nutrient ground truth. The database renders each food across a graded series of portion sizes under otherwise fixed photographic conditions, holding food identity and capture constant while the portion, and thus the nutrient quantity, varies. ASA24 is in this sense a controlled test of portion-size estimation: it isolates whether a model can translate a visible change in portion into a correct change in estimated nutrient amounts.

\paragraph{SNAPMe.} The Surveying Nutrient Assessment with Photographs of Meals (SNAPMe) database, released by the US Department of Agriculture, contains 3,311 food photographs taken by 95 free-living participants on their own mobile phones \cite{larke2023surveying}. Participants photographed every food they consumed and completed parallel ASA24 food records over the same days; each photographed item links through its food record to FNDDS codes and the corresponding 65-nutrient profile. SNAPMe is the first publicly available benchmark of real-life camera-phone food photographs collected for dietary assessment, and its uncontrolled lighting, framing, and clutter make it a direct test of generalization to everyday conditions. ASA24 and SNAPMe are thus complementary: one measures portion-size competence under controlled capture, the other robustness to real-world capture.

\paragraph{NutriBench.} NutriBench is the only evaluation dataset that pairs natural-language meal descriptions with human-verified nutrient labels \cite{dhaliwal2025nutribench}. It comprises 11,857 meal descriptions derived from real-world dietary intake data across eleven countries, drawn from What We Eat in America (the dietary component of NHANES) and the FAO/WHO GIFT database. Each description specifies one or more food items with serving sizes and is labeled with four macronutrients: energy, protein, carbohydrate, and total fat. NutriBench therefore tests text-modality competence at the level of a full meal, including the parsing of multiple items and serving quantities, but its macronutrient-only labels do not probe micronutrient knowledge. It is the realistic meal-description counterpart to the single-food probe provided by FNDDS.

\paragraph{FNDDS.} The Food and Nutrient Database for Dietary Studies (FNDDS), maintained by the US Department of Agriculture, provides the nutrient values for foods reported in What We Eat in America, NHANES, and defines exactly the 65 nutrient and food components used as the label schema throughout this work \cite{bodner2006usda}. As an evaluation dataset it is queried in question-answering form: given the name of a single food, the model must return all 65 nutrient values, scored against the FNDDS entry. Because the query is a single named food, FNDDS removes the perception, meal-parsing, and portion-size confounds present in the other datasets, and is the most direct probe of a model's intrinsic nutrient knowledge.

\subsection{Models}\label{sec:methods-models}

The study compares two classes of model: MLLMs, which are the focus of the paper, and traditional supervised baselines, which test whether an MLLM is necessary for the task at all. Five MLLM families are evaluated. Three are
proprietary (GPT-5, Gemini~3, and Claude Sonnet~4.5) and two are open-weight (the Qwen3-VL family and GLM-4.6V-Flash); the open-weight families additionally serve as the backbones for \textsc{NutriMLLM}. All MLLMs are queried with the same structured prompt (Appendix~\ref{sec:appendix-prompts}), so that differences in performance reflect the model rather than the query.

\paragraph{Proprietary MLLMs.} GPT-5, Gemini~3, and Claude Sonnet~4.5 are evaluated only in their released API-accessible forms as proprietary MLLM baselines. We do not fine-tune these models because their weights are not publicly available, and any vendor-managed tuning, when available, would produce provider-hosted model variants that cannot be released, redistributed, or self-hosted under the same conditions as open-weight models. Therefore, we restrict fine-tuning to open-weight backbones and use proprietary MLLMs to assess the nutrient knowledge already present in frontier general MLLMs.

\paragraph{Open-weight MLLMs.} The open-weight MLLMs are the Qwen3-VL family at 2B/4B/8B/30B parameters and GLM-4.6V-Flash (9B). They play two roles. In their released form they are evaluated as general MLLMs alongside the proprietary models; and because their weights are open, they are also the backbones from which \textsc{NutriMLLM} is fine-tuned. The Qwen3-VL family is included at four parameter scales so that the effect of backbone size can be measured while the architecture and pretraining family are held fixed: the 2B and 4B scales correspond to the on-device regime relevant to privacy-preserving dietary tracking, and the 30B scale to institutional deployment. GLM-4.6V-Flash is included as a second, architecturally distinct open-weight family, so that the fine-tuning recipe is demonstrated on more than one backbone lineage.

\paragraph{NutriMLLM.} \textsc{NutriMLLM} is the family of nutrition-specialized models obtained by fine-tuning each open-weight backbone on the synthetic training corpus described in \S\ref{sec:methods-synth}. One variant is produced per backbone, written as \textsc{NutriMLLM} (Qwen3-VL-30B) and similarly for the other backbones. Fine-tuning uses LoRA for parameter and memory efficiency; the procedure and hyperparameters are given in \S\ref{sec:methods-finetune}. Each variant is evaluated on the same independent datasets and prompt as the general MLLMs, so that it can be compared directly against both its own pre-fine-tune backbone and the proprietary MLLMs.

\paragraph{Supervised baselines.} To test whether comprehensive nutrient estimation requires an MLLM at all, two traditional supervised baselines are included: a text regression model built on a BERT encoder \cite{devlin2019bert} and an image regression model built on a ViT encoder \cite{dosovitskiy2020image}, each with a regression head that outputs nutrient values directly. Both baselines are trained on the same NHANES-derived synthetic corpus used to fine-tune \textsc{NutriMLLM}: the BERT model on the food-text descriptions and the ViT model on the corresponding generated food images. They therefore receive the same supervision as \textsc{NutriMLLM}, so the comparison isolates the contribution of the multimodal language-model backbone, and the large-scale pretraining behind it, rather than confounding it with a difference in training data. Because the regression heads always emit a numeric value and have no mechanism to abstain, these baselines have a NRR of zero by construction, and any unusable predictions they produce are hallucinations rather than abstentions.

\subsection{Synthetic data generation}\label{sec:methods-synth}

The training corpus is built from NHANES 24HRs collected between 2013 and 2023. Each recalled food item carries a set of structured fields, including the food name with its cooking method, the reported portion size, the eating occasion, the time of day, and the food source, together with the complete 65-nutrient FNDDS profile that NHANES assigns to that item. The structured fields are the input from which a food image is generated, and the 65-nutrient profile is the label attached to the resulting image. No new nutrient annotation is produced at any stage; the labels are inherited directly from FNDDS.

\begin{figure}[H]
\centering
\includegraphics[width=\textwidth]{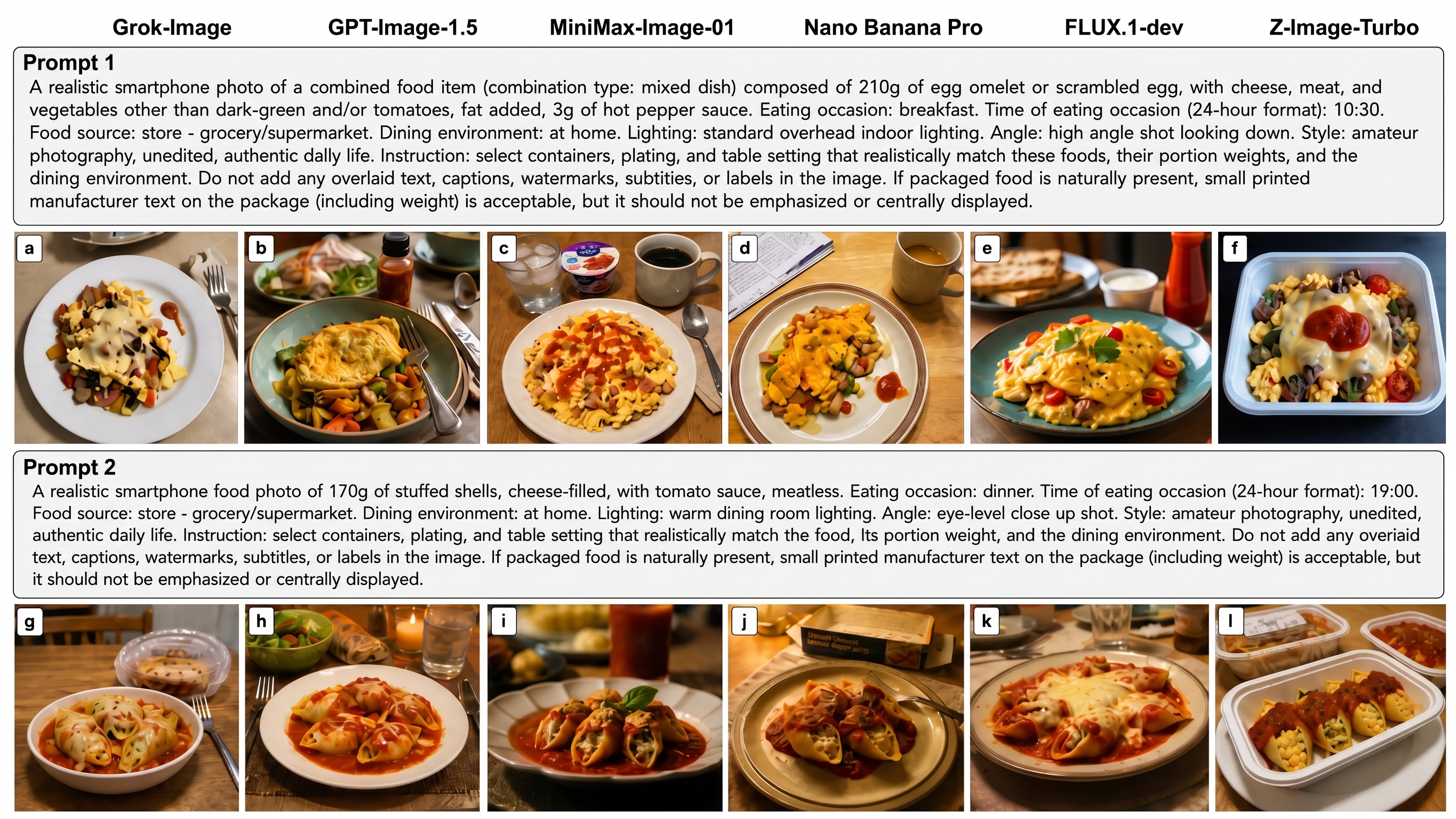}
\caption{\textbf{Comparison of synthetic food images across six text-to-image models.}
Outputs are shown for two NHANES-derived prompts: a breakfast egg-based mixed dish (top row) and stuffed shells for dinner (bottom row). Columns: Grok-Image, GPT-Image-1.5, MiniMax-Image-01, Nano Banana Pro, FLUX.1-dev, and Z-Image-Turbo. FLUX.1-dev and Z-Image-Turbo are open-weight; the remaining four are proprietary. The open-weight pair was used to construct the synthetic training corpus, on the basis of comparable photorealism at no API cost.}\label{fig:generated-images}
\end{figure}

For each food item, the structured recall fields are augmented with simulated photographic variables (lighting, camera angle, and photographic style) and assembled into a descriptive text prompt for a text-to-image model; Fig.~\ref{fig:overview}c illustrates the full process, from a recall to its assembled prompt to the generated image. Each prompt is rendered with two open-weight text-to-image models, Z-Image-Turbo and FLUX.1-dev. As shown in Fig.~\ref{fig:generated-images}, the food images these two models produce are of comparable quality to those of leading proprietary text-to-image models: the depicted food and its portion follow the prompt closely, although the renderings are somewhat less photorealistic. Images from the proprietary models might yield further gains for \textsc{NutriMLLM}, but generating a corpus of this size with them is costly, for example around \$0.134 per image for Nano Banana Pro at 1K--2K resolution (as of April 2026), whereas Z-Image-Turbo and FLUX.1-dev incur no API cost. Rendering every prompt with both open-weight generators, each of which imposes a distinct visual style (Fig.~\ref{fig:generated-images}), also keeps the corpus from inheriting the stylistic signature of any single generator. The pipeline yields a synthetic multimodal corpus of around 1.1~million image--description--nutrient triplets, each pairing a generated food image with a text description and its 65-nutrient label.

Using two generators thus removes dependence on any single image model, broadens the visual distribution the model is trained on, and avoids the per-image cost that proprietary generators would impose at this scale. The ablation in \S\ref{sec:results-ablations} shows that nutrient-knowledge transfer is robust to the choice of generator, with the union of both performing best.

\subsection{Domain adaptation via LoRA fine-tuning}\label{sec:methods-finetune}

Each \textsc{NutriMLLM} variant is obtained by fine-tuning one open-weight backbone (Qwen3-VL at 2B/4B/8B/30B, and GLM-4.6V-Flash at 9B) on the synthetic corpus of \S\ref{sec:methods-synth}. Fine-tuning spans both input modalities, matching the two on which the model is evaluated: an image example presents the generated food image, while a text example presents the food's name and portion size as a short textual description. In both cases the input is paired with the nutrient-estimation prompt (Prompt~2, Appendix~\ref{sec:appendix-prompts}). The image-generation template (Prompt~1) is used only upstream in \S\ref{sec:methods-synth} to render the image and is never shown to the model at this stage.

The supervision target is the full step-by-step response that Prompt~2 elicits: a brief reasoning that first identifies the food and then states its portion, followed by the complete 65-nutrient profile as a JSON object. For image inputs, the food identity and portion in this target are quantities the model must learn to infer from the image; for text inputs they restate the given description, and the nutrient profile is the new prediction. Crucially, only the image is synthetic; the target itself is real data. Every element of it (the food name, the portion, and the 65 nutrient values) is taken directly from the NHANES recalls, so the reasoning trace is built entirely from ground truth, with no teacher model and no annotation beyond what the recall already provides. Every label is complete: all 65 values are populated for every item, so the placeholder \texttt{na} never appears in a target and the model is never trained to omit a value. Fine-tuning optimizes the standard supervised objective on this target using LoRA \cite{hu2022lora} for parameter-efficient adaptation.

\subsection{Evaluation protocol}\label{sec:methods-eval}

All models are evaluated on the four independent datasets of \S\ref{sec:methods-data} under a common protocol. Every model receives the same input for a given query: the image datasets (ASA24, SNAPMe) are evaluated with the multimodal prompt and the text datasets (NutriBench, FNDDS) with its text-only counterpart (Prompt~2, Appendix~\ref{sec:appendix-prompts}). For each query the model is asked to return a value for every nutrient in the dataset's label set, and is permitted to return the placeholder \texttt{na} for any nutrient it will not estimate. Proprietary MLLMs are queried through their vendor APIs, and open-weight models are run locally, but the prompt, the requested output format, and the sampled evaluation items are identical across all models. Because every training label is complete (\S\ref{sec:methods-finetune}), \textsc{NutriMLLM} is never trained to abstain; any \texttt{na} it returns at inference is a response to the prompt's option to abstain rather than a behavior taught by the training labels. Model behavior is scored with the four-component framework introduced in \S\ref{sec:results-validation}: NRR, HR, UPR, and accuracy among answered predictions, reported as abstention-adjusted $\mathrm{SMAPE}^{\mathrm{adj}}$, with formal definitions given in Appendix~\ref{sec:eval_metrics}.

\section{Discussion}\label{sec:discussion}

Our main finding is a diagnosis. Testing general MLLMs on text descriptions as well as on images showed that their failure on micronutrients is not caused by weak vision: a model that cannot estimate a vitamin or fatty acid from a photograph also cannot estimate it from the food's written name. The problem is missing nutrient knowledge, not poor perception, and better image encoders would not fix it. The likely cause is a lack of the right data. Comprehensive nutrient profiles are rare in these models' training data: recipes and nutrition labels usually give energy and the macronutrients, but seldom the complete set of vitamins, minerals, and fatty acids. And no large-scale food dataset has paired images with comprehensive nutrient labels, so the knowledge has had no source to learn from. The gap is therefore structural, set by the available data rather than by model size: the failure on micronutrients persists across the full size spectrum, from the smallest open-weight models to the largest proprietary ones, even though greater scale does improve overall accuracy. Scaling general models further will not close it; the gap closes only when the missing data is supplied.

\textsc{NutriMLLM} closes this gap. Fine-tuning on the synthetic corpus sharply reduces both abstention and hallucination across the 65 nutrients. This is a real gain, not just the model being pushed to answer more often: the ablations in \S\ref{sec:results-ablations} show that the nutrients it now answers also become more accurate, with a clear leftward shift in $\mathrm{SMAPE}^{\mathrm{adj}}$. The model answers more, and it answers better.

The recipe works because of two facts. First, the expensive half of the supervision already exists. The field has treated the bottleneck as the cost of labeling food images with full nutrient profiles, but national surveys have linked structured dietary recalls to complete profiles for decades. What is missing is not the label but the image, and the image is the cheap part to create: rather than label images with nutrients, we generate images for records that are already labeled. Only the image is synthetic; the labels are real survey data. Second, nutrient estimation depends mainly on what an image shows, the food and its portion, not on how photorealistic it is. Current text-to-image models get the food and portion right well enough that this signal survives, even when the images are not perfectly realistic (Fig.~\ref{fig:generated-images}). The transfer to the real photographs of SNAPMe confirms it: even though every image \textsc{NutriMLLM} trained on was generated, it generalizes to real photographs it never saw, because what it learned is nutrient knowledge, not the look of any one generator.

This points to the paper's main contribution. The most visible result is \textsc{NutriMLLM} itself, but the more lasting one is the recipe that produced it. The recipe does not depend on the backbone it adapts, so it is not limited to the open-weight models used here: developers of proprietary MLLMs could apply the same procedure to their own models to add the nutrient knowledge they currently lack. The paper therefore provides a reusable method for building nutrition-specialized models on any backbone, not just a single model. The result also carries a broader lesson: for narrow, knowledge-heavy problems, reliable performance comes from building the right specialized supervision, not from making general models larger. Because the models are open-weight and small, they can be run locally and inexpensively, so meal images need not be sent to an external service and use does not depend on a vendor API. Whether the same approach, turning structured records into multimodal supervision through generation, works in other clinical domains where records exist without paired images is left for future work.

Several limitations remain. The evaluation is retrospective and based on benchmarks: \textsc{NutriMLLM} has not been tested in a real clinical workflow, and its estimates have not been compared against biomarkers of nutrient status, so the results show technical capability, not clinical readiness. The training images are synthetic; although their nutrient-relevant content transfers to real photographs, they may lack appearance cues that real food carries, and the ground-truth values are themselves FNDDS estimates rather than direct laboratory measurements. Portion size is inferred from a single image, which is inherently ambiguous, and one view cannot fully resolve occluded or stacked food. Finally, two design questions are left for future ablation: how much each structured recall field adds beyond food name and portion, and how performance scales with the size of the generated corpus.

One clear next step follows from a limitation of our data. The recalls and labels come entirely from US sources, NHANES and FNDDS, so the food distribution reflects one country's diet, and \textsc{NutriMLLM} is likely strongest on foods common in the United States. This is not a limitation of the method. The recipe needs only structured recall fields linked to a nutrient profile, and many countries maintain exactly these resources: national 24-hour-recall surveys such as the UK National Diet and Nutrition Survey \cite{venables2022data}, the Japan National Health and Nutrition Survey \cite{ikeda2015data}, and the German National Nutrition Survey \cite{heuer2015food}, each paired with a national food-composition database like FNDDS, such as McCance and Widdowson's The Composition of Foods Integrated Dataset \cite{mccance2014mccance} in the UK and the tables harmonized across Europe through EuroFIR \cite{finglas2014assessing}. Applying the recipe to these sources would give country-specific variants, and pooling images generated from several countries' surveys into one corpus would produce a worldwide \textsc{NutriMLLM} covering many cuisines rather than a single national diet.

A second direction concerns the range of tasks. This paper tests \textsc{NutriMLLM} on one capability, estimating a full nutrient profile from a single food image. That capability is a prerequisite for the applications that motivate the work, but it is not the application itself. The broader goal in Fig.~\ref{fig:overview}a is a system that supports dietary assessment from the many ways people record meals, including video and speech as well as photographs, and that turns reliable nutrient estimates into personalized guidance and population-scale surveillance. \textsc{NutriMLLM} provides the nutrient knowledge these tasks need; building it into each application and evaluating it there is what would turn a reliable estimator into a deployed tool, and we see that as the path from this work to clinical and public-health impact.

\section{Conclusion}\label{sec:conclusion}

We showed that existing MLLMs, including the leading proprietary ones, do not yet have reliable comprehensive nutrient knowledge, that this failure falls on the micronutrients that matter most clinically, and that it persists even without an image, which places the deficit in nutrient knowledge rather than visual perception. To close the gap, we turned a decade of national-survey 24HRs into a synthetic multimodal corpus of around 1.1~million image–description–nutrient triplets, and fine-tuned open-weight backbones on this corpus to produce \textsc{NutriMLLM}, a family of nutrition-specialized vision-language models. On real food images, \textsc{NutriMLLM} reaches broad coverage across all 65 nutrients, and its largest variant matches or exceeds the proprietary MLLMs on $\mathrm{SMAPE}^{\mathrm{adj}}$ for most nutrients, while staying small enough to run on-device. Beyond the models, the main contribution is a reusable recipe: because it needs only structured survey records, it can be applied to any backbone, including proprietary ones, and extended from the US sources used here to the national surveys of other countries. We release the evaluation framework, the generated dataset, and the \textsc{NutriMLLM} weights so that others can build on them. The necessary next step is to establish whether this technical capability translates into clinical and public-health benefit, through prospective evaluation and integration into real dietary-assessment workflows.

\backmatter












\begin{appendices}

\section{Evaluation Framework}\label{sec:eval_metrics}
 
\subsection{Conceptual Rationale}
 
Large language models performing quantitative nutrition prediction
exhibit two qualitatively distinct failure modes in addition to ordinary estimation error:
 
\paragraph{Abstention.}
The model does not provide a numeric estimate. This typically reflects uncertainty, insufficient domain knowledge, or inability to extract relevant information from the input. In this case, the model effectively declines to answer.
 
\paragraph{Hallucination.}
The model provides a numeric estimate, but the value is statistically implausible relative to observed nutritional distributions. This reflects overconfident reasoning breakdown, where the model outputs an extreme or unrealistic value despite insufficient grounding. These behaviors are fundamentally different. Abstention reflects uncertainty or knowledge limitation; hallucination reflects confident but incorrect estimation. Traditional accuracy metrics alone cannot distinguish these failure modes.
 
Model performance is therefore evaluated along three complementary dimensions:
 
\begin{enumerate}
\item Non-response behavior (abstention),
\item Hallucination behavior (implausible predictions),
\item Numerical accuracy among answered predictions.
\end{enumerate}
 
All metrics are computed at the nutrient level.
 
Let $N$ denote the total number of samples. Let $t_{ij}$ and $p_{ij}$ denote the ground-truth and predicted values, respectively, for sample $i$ and nutrient $j$. Negative predictions are treated as missing.
 
\subsection{Non-Response and Hallucination Metrics}
 
\subsubsection{Non-Response Rate (NRR)}
 
A prediction is considered missing if no valid numeric value is produced.
 
\[
\NRR_j
=
\frac{1}{N}
\sum_{i=1}^{N}
\mathbf{1}[p_{ij} \text{ is missing}]
\times 100.
\]
 
NRR quantifies the proportion of samples for which the model abstains.
 
\subsubsection{Hallucination Rate (HR)}
 
Hallucinations are defined relative to the empirical ground-truth distribution.
 
For each nutrient $j$, let
 
\[
\tau_j = Q_{1-\alpha}(t_{\cdot j}),
\]
 
where $Q_{1-\alpha}$ denotes the $(1-\alpha)$ quantile of the ground-truth distribution (e.g., $\alpha = 0.005$ corresponding to the 99.5th percentile).
 
A prediction is considered hallucinatory if:
 
\[
p_{ij} > \tau_j.
\]
 
The Hallucination Rate (HR) is defined as:
 
\[
\HR_j
=
\frac{1}{N}
\sum_{i=1}^{N}
\mathbf{1}[p_{ij} \text{ predicted and } p_{ij} > \tau_j]
\times 100.
\]
 
This quantile-based definition anchors hallucination detection to observed nutritional reality rather than to the model's own prediction distribution. It is robust to skewed or multi-modal nutrient distributions.
 
\subsubsection{Unusable Prediction Rate (UPR)}
 
We define the Unusable Prediction Rate as:
 
\[
\UPR_j = \NRR_j + \HR_j.
\]
 
UPR represents the overall proportion of nutritionally unusable outputs,
capturing both abstention and implausible predictions.
 
\subsection{Predictive Accuracy Metrics}

Accuracy is measured with the Symmetric Mean Absolute Percentage Error (SMAPE), which suits this task for two reasons: the 65 nutrients span very different ranges and units, from kilocalories to micrograms, so absolute errors cannot be averaged across them, and many ground-truth values are zero, which leaves the ordinary percentage error (MAPE) undefined. SMAPE handles both and is bounded between 0 and 200\%, so a hallucinated extreme value cannot dominate the average. We report an abstention-adjusted SMAPE, which scores every ground-truth value and penalizes each missing prediction, so that a model cannot inflate its accuracy by abstaining on the difficult nutrients.

For a single answered prediction, the symmetric absolute percentage error is

\[
\SMAPE(p_{ij}, t_{ij})
=
\frac{2\,|p_{ij} - t_{ij}|}
{|p_{ij}| + |t_{ij}| + \varepsilon}
\times 100,
\]
 
\subsubsection{Covered Metrics}
 
The covered SMAPE evaluates accuracy on answered predictions only. Let

\[
S_j = \{ i : t_{ij} \text{ known and } p_{ij} \text{ present} \}.
\]

\[
\SMAPEcov_j
=
\frac{1}{|S_j|}
\sum_{i \in S_j}
\SMAPE(p_{ij}, t_{ij}).
\]

When $p_{ij}=0$ and $t_{ij}=0$, the contribution is defined as 0. Hallucinated predictions are not excluded and therefore increase the error.
 
\subsubsection{Abstention-Adjusted Metrics}
 
The abstention-adjusted SMAPE evaluates accuracy across all ground-truth values, not only the answered ones. Let

\[
V_j = \{ i : t_{ij} \text{ known} \}.
\]

For each missing prediction we assign a penalty equal to the $Q_{0.9}$ quantile of the model's observed errors for that nutrient (falling back to a dataset-level baseline if too few answered samples exist):

\[
\SMAPEadj_j
=
\frac{1}{|V_j|}
\sum_{i \in V_j}
\begin{cases}
\SMAPE(p_{ij}, t_{ij}), & p_{ij} \text{ present}, \\
\mathcal{P}^{\mathrm{SMAPE}}_j, & p_{ij} \text{ missing}.
\end{cases}
\]
 
\subsection{Complementarity of Metrics}
 
Together, NRR, HR, UPR, and SMAPE capture complementary aspects of model behavior: knowledge limitation, overconfident implausible estimation, overall usability, and numerical accuracy (the last reported as covered SMAPE on answered cases and $\mathrm{SMAPE}^{\mathrm{adj}}$ across all cases). Reporting them together, rather than accuracy alone, prevents the misleading conclusions a single metric can produce and lets models that differ in calibration, robustness, and selective answering be compared on a common footing.

\section{Results on Text Food-Description Datasets}\label{sec:appendix-text-results}

This appendix provides the per-nutrient figures and the NutriBench comparison table summarized in the main text. Both benchmarks evaluate models on food-text inputs only and therefore isolate the nutrient-knowledge component of MLLM performance from any visual-perception component.

\subsection{FNDDS}

\begin{figure}[H]
\centering
\includegraphics[width=0.96\textwidth]{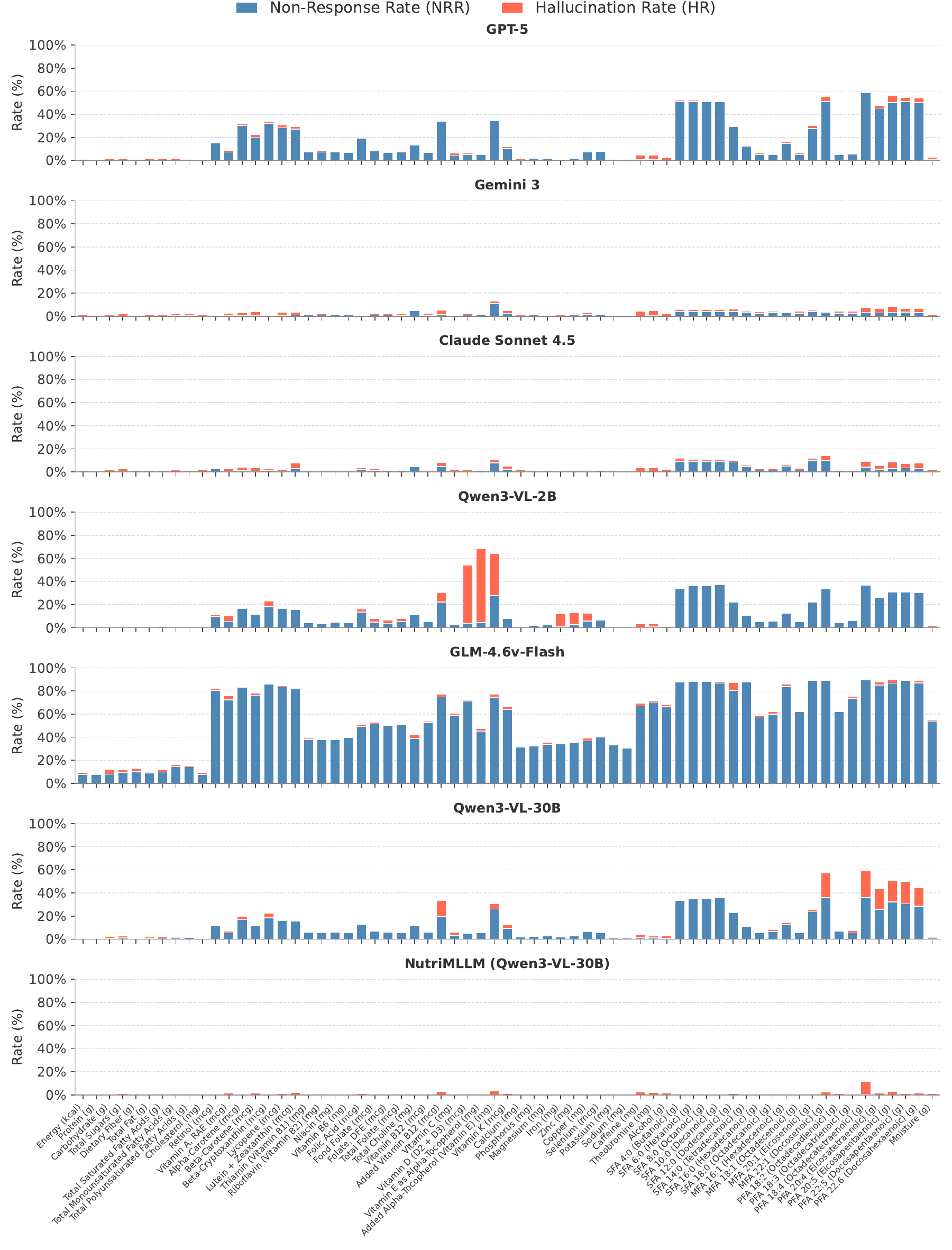}
\caption{\textbf{Non-response and hallucination rates on the FNDDS text-only benchmark.}
Per-nutrient NRR and HR, evaluated on food descriptions rather than images, with model groupings as in Fig.~\ref{fig:metric12-asa24}. Failure on micronutrients persists in the absence of any visual input, indicating that the deficit in general MLLMs is one of nutrient knowledge rather than visual perception.}\label{fig:metric12-fndds}
\end{figure}

\begin{figure}[H]
\centering
\includegraphics[width=\textwidth]{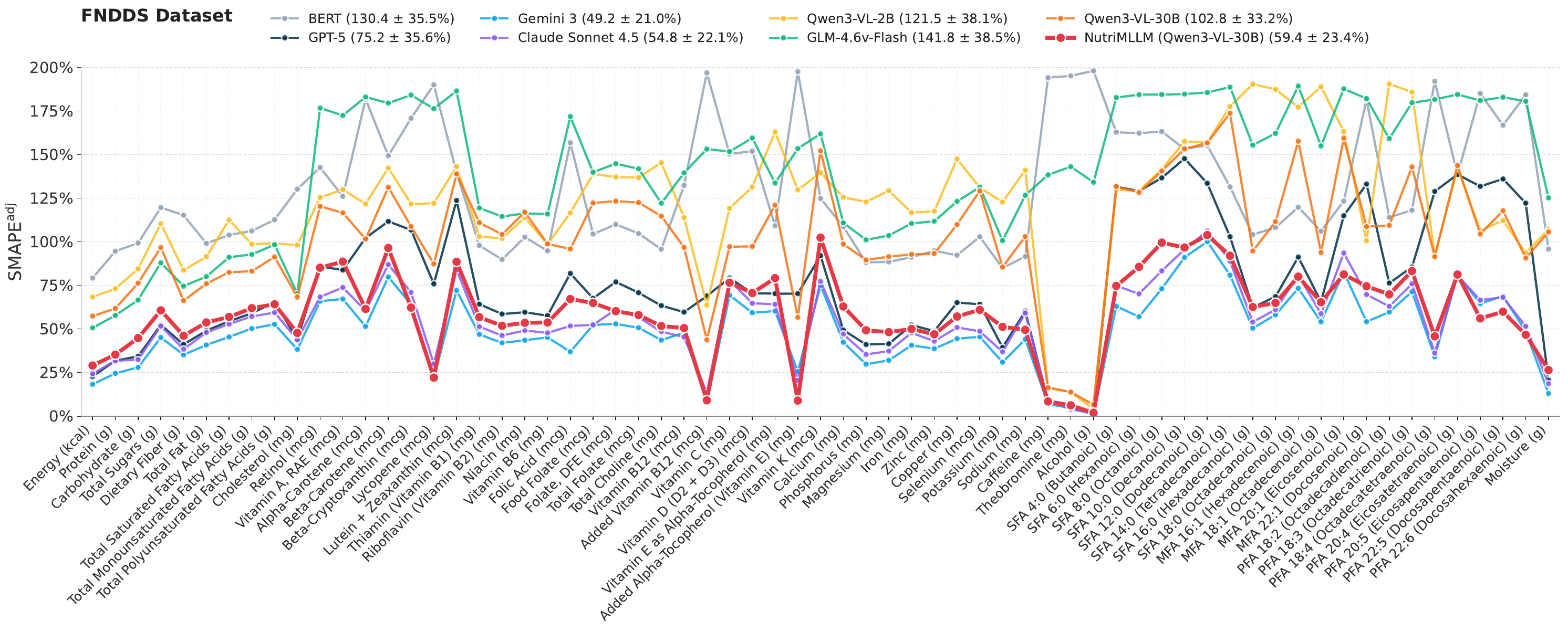}
\caption{\textbf{Per-nutrient accuracy on FNDDS.}
Covered (top) and abstention-adjusted (bottom) SMAPE on the text-only FNDDS benchmark, evaluated and grouped as in Fig.~\ref{fig:metric3-asa24-SNAPMe}. \textsc{NutriMLLM} closes the gap to proprietary baselines across all 65 nutrients on text inputs, with the largest gains again concentrated on micronutrients.}\label{fig:metric3-fndds}
\end{figure}

\subsection{NutriBench}

\begin{sidewaystable}
\caption{\textbf{Comparison of \textsc{NutriMLLM} against baseline approaches on the four target nutrients of NutriBench.}
UPR is the Unusable Prediction Rate (NRR\,$+$\,HR); $\mathrm{SMAPE}^{\mathrm{adj}}$ is the abstention-adjusted Symmetric Mean Absolute Percentage Error. Lower is better for both metrics. \textbf{Bold} marks the best value in each column overall; \underline{underline} marks the best value among non-proprietary models. Both the supervised BERT baseline and the \textsc{NutriMLLM} variants were trained on the same NHANES-derived synthetic corpus and never saw NutriBench during training, so both are evaluated zero-shot on this benchmark. \textsc{NutriMLLM} (Qwen3-VL-30B) achieves the lowest UPR among non-proprietary models on every nutrient and the lowest $\mathrm{SMAPE}^{\mathrm{adj}}$ among non-proprietary models on three of four nutrients, despite being a generalist 65-nutrient predictor rather than a model specialized for these four macronutrients.}\label{tab:nutrimllm-results}
\begin{tabular*}{\textheight}{@{\extracolsep\fill}lcccccccc}
\toprule%
& \multicolumn{4}{@{}c@{}}{UPR (\%) $\downarrow$} & \multicolumn{4}{@{}c@{}}{$\mathrm{SMAPE}^{\mathrm{adj}}$ (\%) $\downarrow$} \\\cmidrule{2-5}\cmidrule{6-9}%
Model & Energy (kcal) & Protein (g) & Carb. (g) & Total Fat (g) & Energy (kcal) & Protein (g) & Carb. (g) & Total Fat (g) \\
\midrule
\multicolumn{9}{@{}l}{\emph{Supervised Baseline}} \\
\quad BERT (+ regression head) & 0.1 & 4.5 & 0.5 & 2.0 & 56.6 & 77.5 & 65.0 & 76.1 \\
\midrule
\multicolumn{9}{@{}l}{\emph{Open-Weight MLLM Baselines}} \\
\quad Qwen3-VL-2B    & 39.0 & 39.2 & 37.5 & 39.1 & 95.0  & 106.1 & 97.8  & 111.0 \\
\quad Qwen3-VL-4B    & 23.6 & 24.0 & 18.0 & 24.6 & 108.4 & 114.3 & 117.1 & 124.0 \\
\quad Qwen3-VL-8B    & 6.6  & 6.6  & 5.9  & 6.7  & 56.3  & 75.2  & 66.7  & 87.3  \\
\quad GLM-4.6V-Flash & 25.5 & 25.2 & 25.4 & 26.8 & 75.0  & 82.5  & 81.2  & 95.2  \\
\midrule
\multicolumn{9}{@{}l}{\emph{Proprietary MLLM Baselines}} \\
\quad Claude Sonnet~4.5 & 0.2 & 0.4 & 0.2 & \textbf{0.2} & 34.1 & 42.2 & 36.7 & 52.9 \\
\quad Gemini~3          & 0.7 & 0.8 & 0.6 & 0.8 & \textbf{25.3} & 34.4 & \textbf{29.1} & \textbf{41.8} \\
\quad GPT-5             & 0.1 & \textbf{0.3} & \textbf{0.1} & \textbf{0.2} & 32.1 & \textbf{32.4} & 37.3 & 50.5 \\
\midrule
\multicolumn{9}{@{}l}{\emph{\textsc{NutriMLLM} (Ours)}} \\
\quad Qwen3-VL-2B    & 2.8 & 2.8 & 4.4 & 2.6 & 128.4 & 100.7 & 74.5 & 105.9 \\
\quad Qwen3-VL-4B    & 1.4 & 1.7 & 1.3 & 1.5 & 70.4  & 61.1  & 63.0 & 78.0  \\
\quad Qwen3-VL-8B    & 0.5 & 0.6 & 0.5 & 0.8 & 53.8  & \underline{43.8} & 52.9 & 71.0  \\
\quad Qwen3-VL-30B   & \textbf{\underline{0.0}} & \underline{0.4} & \underline{0.4} & \underline{0.3} & \underline{47.4} & 48.9 & \underline{46.3} & \underline{54.8} \\
\quad GLM-4.6V-Flash & 0.9 & 1.0 & 2.1 & 1.2 & 58.9  & 54.7  & 51.4 & 62.7  \\
\botrule
\end{tabular*}
\end{sidewaystable}

\section{Training Dynamics}\label{sec:appendix-training-dynamics}

\begin{figure}[H]
\centering
\includegraphics[width=\textwidth]{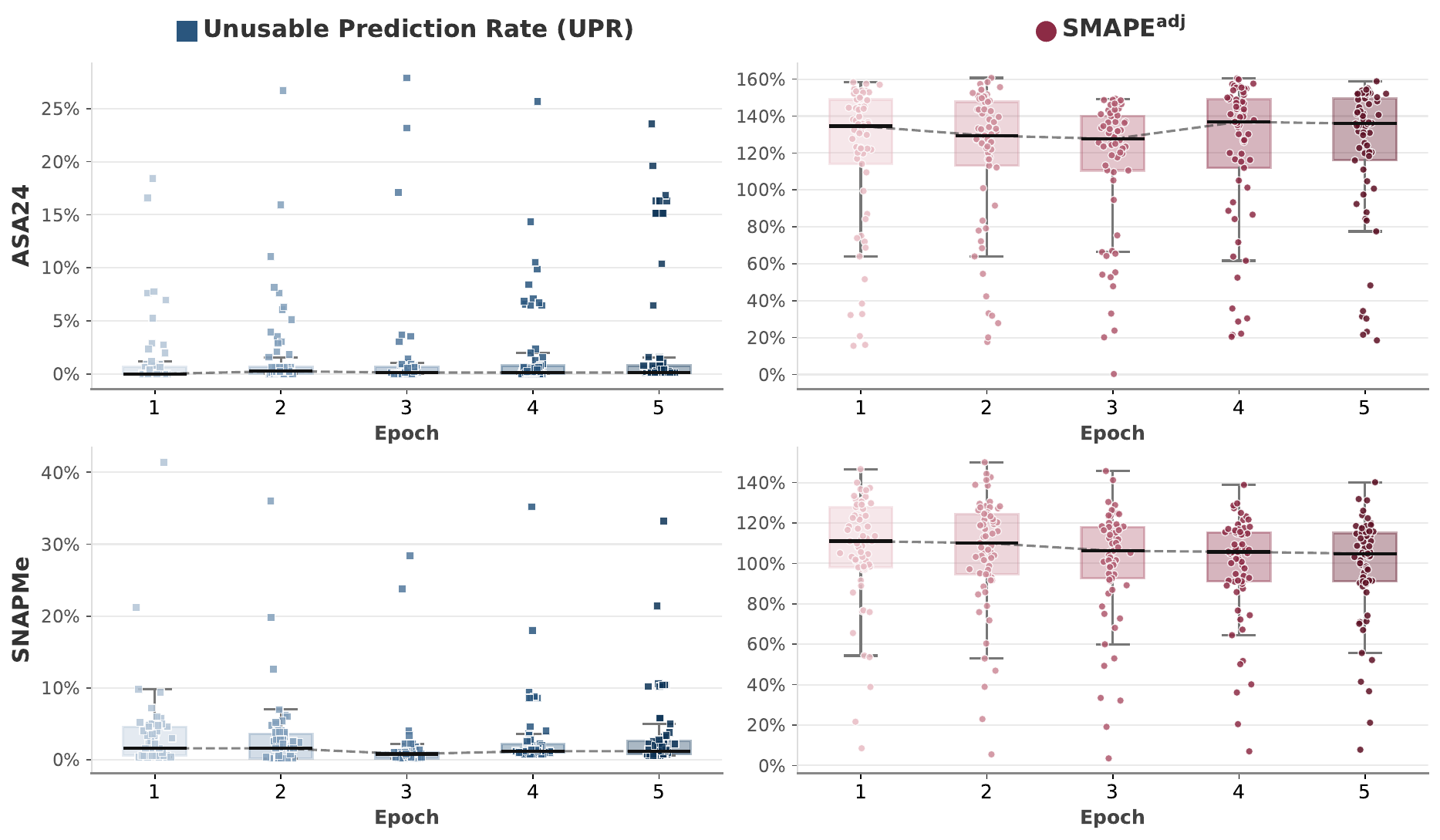}
\caption{\textbf{Validation performance across \textsc{NutriMLLM} training checkpoints.}
UPR (left) and $\mathrm{SMAPE}^{\mathrm{adj}}$ (right) on the two independent datasets (ASA24 and SNAPMe) as a function of training epoch. UPR is low from the first epoch and stays low throughout, while $\mathrm{SMAPE}^{\mathrm{adj}}$ improves over the first few epochs and plateaus by around epoch~3.}\label{fig:checkpoint-comparison}
\end{figure}

\section{Training Implementation Details}\label{sec:appendix-training}

Fine-tuning was carried out with the LLaMA-Factory framework \cite{zheng2024llamafactory} on four H200 GPUs. Upon publication, we will make publicly available the training and evaluation code, LLaMA-Factory configuration files, fine-tuned \textsc{NutriMLLM} weights, and the synthetic multimodal corpus of approximately 1.1 million image–description–nutrient triplets, subject to the licenses of the underlying data sources, base models, and image-generation models. If redistribution of any generated images is restricted by third-party model terms, we will release the corresponding prompts, nutrient labels, metadata, and generation scripts needed to reproduce the corpus.

\section{Prompt Templates}\label{sec:appendix-prompts}

\begin{figure}[H]
\centering
\includegraphics[width=\textwidth]{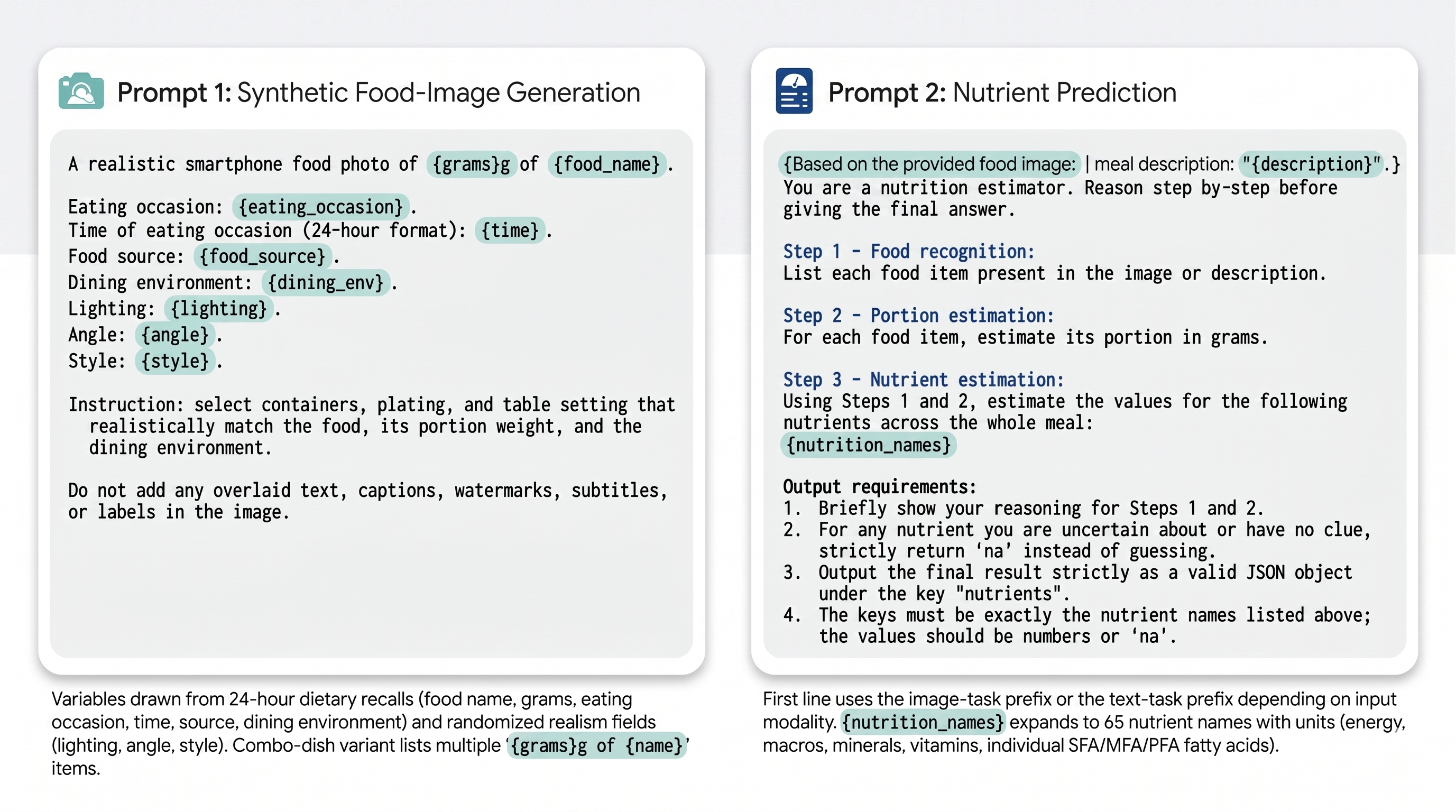}
\caption{\textbf{Prompt templates used in \textsc{NutriMLLM}.}
\textbf{Prompt~1} assembles structured 24-hour-recall fields (food name, cooking method, portion-size descriptor, eating occasion, time of day, food source) together with simulated photographic variables (lighting condition, camera angle, photography style, table setting) into a natural-language description for text-to-image generation.
\textbf{Prompt~2} queries the model for 65 nutrient values from a food image or text description, requiring a strict JSON object in which the placeholder \texttt{na} is permitted for entries the model declines to estimate. Curly braces denote runtime-filled variables.}\label{fig:prompts}
\end{figure}

Figure~\ref{fig:prompts} reproduces the two prompt templates used throughout this work: Prompt~1 for synthetic image generation and Prompt~2 for nutrient estimation.

\end{appendices}

\newpage

\bibliography{sn-bibliography}

@inproceedings{lozoff2006iron,
  title={Iron deficiency and brain development},
  author={Lozoff, Betsy and Georgieff, Michael K},
  booktitle={Seminars in pediatric neurology},
  volume={13},
  number={3},
  pages={158--165},
  year={2006},
  organization={Elsevier}
}

@article{holick2007vitamin,
  title={Vitamin D deficiency},
  author={Holick, Michael F},
  journal={New England journal of medicine},
  volume={357},
  number={3},
  pages={266--281},
  year={2007},
  publisher={Mass Medical Soc}
}

@article{mrc1991prevention,
  title={Prevention of neural tube defects: results of the Medical Research Council Vitamin Study},
  author={MRC Vitamin Study Research Group and others},
  journal={The lancet},
  volume={338},
  number={8760},
  pages={131--137},
  year={1991},
  publisher={Elsevier}
}

@article{zimmermann2008iodine,
  title={Iodine-deficiency disorders},
  author={Zimmermann, Michael B and Jooste, Pieter L and Pandav, Chandrakant S},
  journal={The Lancet},
  volume={372},
  number={9645},
  pages={1251--1262},
  year={2008},
  publisher={Elsevier}
}

@article{stabler2013vitamin,
  title={Vitamin B12 deficiency},
  author={Stabler, Sally P},
  journal={New England Journal of Medicine},
  volume={368},
  number={21},
  pages={2041--2042},
  year={2013}
}

@article{sommer1983increased,
  title={Increased mortality in children with mild vitamin A deficiency},
  author={Sommer, Alfred and Hussaini, Gusti and Tarwotjo, Ignatius and Susanto, Djoko},
  journal={The Lancet},
  volume={322},
  number={8350},
  pages={585--588},
  year={1983},
  publisher={Elsevier}
}

@misc{prasad2003zinc,
  title={Zinc deficiency: Has been known of for 40 years but ignored by global health organisations},
  author={Prasad, Ananda S},
  journal={Bmj},
  volume={326},
  number={7386},
  pages={409--410},
  year={2003},
  publisher={British Medical Journal Publishing Group}
}

@article{weaver2016calcium,
  title={Calcium plus vitamin D supplementation and risk of fractures: an updated meta-analysis from the National Osteoporosis Foundation},
  author={Weaver, Connie M and Alexander, Dominik D and Boushey, Carol J and Dawson-Hughes, Bess and Lappe, Joan M and LeBoff, Meryl S and Liu, Simin and Looker, Anne C and Wallace, TC and Wang, DD},
  journal={Osteoporosis International},
  volume={27},
  number={1},
  pages={367--376},
  year={2016},
  publisher={Springer}
}

@article{thompson2017dietary,
  title={Dietary assessment methodology},
  author={Thompson, Frances E and Subar, Amy F},
  journal={Nutrition in the Prevention and Treatment of Disease},
  pages={5--48},
  year={2017},
  publisher={Elsevier}
}

@article{lo2024dietary,
  title={Dietary assessment with multimodal ChatGPT: A systematic analysis},
  author={Lo, Frank P-W and Qiu, Jianing and Wang, Zeyu and Chen, Junhong and Xiao, Bo and Yuan, Wu and Giannarou, Stamatia and Frost, Gary and Lo, Benny},
  journal={IEEE Journal of Biomedical and Health Informatics},
  volume={28},
  number={12},
  pages={7577--7587},
  year={2024},
  publisher={IEEE}
}

@inproceedings{khamesian2025nutrigen,
  title={NutriGen: Personalized meal plan generator leveraging large language models to enhance dietary and nutritional adherence},
  author={Khamesian, Saman and Arefeen, Asiful and Carpenter, Stephanie M and Ghasemzadeh, Hassan},
  booktitle={2025 47th Annual International Conference of the IEEE Engineering in Medicine and Biology Society (EMBC)},
  pages={1--7},
  year={2025},
  organization={IEEE}
}

@article{carrillo2025llms,
  title={LLMs for energy and macronutrients estimation using only text data from 24-hour dietary recalls: a parameter-efficient fine-tuning experiment using a 10-shot prompt},
  author={Carrillo-Larco, Rodrigo M},
  journal={arXiv preprint arXiv:2509.13268},
  year={2025}
}

@article{gjorgjevikj2026large,
  title={Large language models in food and nutrition science: Opportunities, challenges, and the case of FoodyLLM},
  author={Gjorgjevikj, Ana and Martinc, Matej and Cenikj, Gjorgjina and Stojanov, Riste and Drole, Jan and Ispirova, Gordana and Menichetti, Giulia and Ogrinc, Nives and Trajanov, Dimitar and D{\v{z}}eroski, Sa{\v{s}}o and others},
  journal={Current research in food science},
  volume={12},
  pages={101351},
  year={2026},
  publisher={Elsevier}
}

@inproceedings{
dhaliwal2025nutribench,
title={NutriBench: A Dataset for Evaluating Large Language Models in Nutrition Estimation from Meal Descriptions},
author={Mehak Preet Dhaliwal and Andong Hua and Laya Pullela and Ryan Burke and Yao Qin},
booktitle={The Thirteenth International Conference on Learning Representations},
year={2025},
url={https://openreview.net/forum?id=6LtdZCyuZR}
}

@inproceedings{bossard2014food,
  title={Food-101--mining discriminative components with random forests},
  author={Bossard, Lukas and Guillaumin, Matthieu and Van Gool, Luc},
  booktitle={European conference on computer vision},
  pages={446--461},
  year={2014},
  organization={Springer}
}

@article{marin2021recipe1m+,
  title={Recipe1m+: A dataset for learning cross-modal embeddings for cooking recipes and food images},
  author={Mar{\i}n, Javier and Biswas, Aritro and Ofli, Ferda and Hynes, Nicholas and Salvador, Amaia and Aytar, Yusuf and Weber, Ingmar and Torralba, Antonio},
  journal={IEEE Transactions on Pattern Analysis and Machine Intelligence},
  volume={43},
  number={1},
  pages={187--203},
  year={2021}
}

@inproceedings{thames2021nutrition5k,
  title={Nutrition5k: Towards automatic nutritional understanding of generic food},
  author={Thames, Quin and Karpur, Arjun and Norris, Wade and Xia, Fangting and Panait, Liviu and Weyand, Tobias and Sim, Jack},
  booktitle={Proceedings of the IEEE/CVF conference on computer vision and pattern recognition},
  pages={8903--8911},
  year={2021}
}

@article{johnson2013national,
  title={National health and nutrition examination survey. Analytic guidelines, 1999-2010},
  author={Johnson, Clifford L and Paulose-Ram, Ryne and Ogden, Cynthia L and Carroll, Margaret D and Kruszan-Moran, Deanna and Dohrmann, Sylvia M and Curtin, Lester R},
  year={2013}
}

@article{bodner2006usda,
  title={USDA food and nutrient database for dietary studies: released on the web},
  author={Bodner-Montville, Janice and Ahuja, Jaspreet KC and Ingwersen, Linda A and Haggerty, Etta Susanne and Enns, Cecilia Wilkinson and Perloff, Betty P},
  journal={Journal of food composition and analysis},
  volume={19},
  pages={S100--S107},
  year={2006},
  publisher={Elsevier}
}

@article{yan2025dietai24,
  title={DietAI24 as a framework for comprehensive nutrition estimation using multimodal large language models},
  author={Yan, Runze and Luo, Hanqi and Lu, Jiaying and Liu, Darren and Posluszny, Hannah and Dhaliwal, Mehak Preet and MacLeod, Janice and Qin, Yao and Yang, Carl and Hartman, Terry J and others},
  journal={Communications Medicine},
  volume={5},
  number={1},
  pages={458},
  year={2025},
  publisher={Nature Publishing Group UK London}
}

@misc{asa24portionimages,
  author       = {{National Cancer Institute}},
  title        = {{ASA24 Portion Size Image Database}},
  howpublished = {Epidemiology and Genomics Research Program, National Cancer Institute},
  note         = {\url{https://epi.grants.cancer.gov/asa24/resources/portionsize.html}}
}

@article{larke2023surveying,
  title={Surveying nutrient assessment with photographs of meals (SNAPMe): a benchmark dataset of food photos for dietary assessment},
  author={Larke, Jules A and Chin, Elizabeth L and Bouzid, Yasmine Y and Nguyen, Tu and Vainberg, Yael and Lee, Dong Hee and Pirsiavash, Hamed and Smilowitz, Jennifer T and Lemay, Danielle G},
  journal={Nutrients},
  volume={15},
  number={23},
  pages={4972},
  year={2023},
  publisher={MDPI}
}

@article{bai2025qwen3,
  title={Qwen3-vl technical report},
  author={Bai, Shuai and Cai, Yuxuan and Chen, Ruizhe and Chen, Keqin and Chen, Xionghui and Cheng, Zesen and Deng, Lianghao and Ding, Wei and Gao, Chang and Ge, Chunjiang and others},
  journal={arXiv preprint arXiv:2511.21631},
  year={2025}
}

@article{hong2025glm,
  title={Glm-4.5 v and glm-4.1 v-thinking: Towards versatile multimodal reasoning with scalable reinforcement learning},
  author={Hong, Wenyi and Yu, Wenmeng and Gu, Xiaotao and Wang, Guo and Gan, Guobing and Tang, Haomiao and Cheng, Jiale and Qi, Ji and Ji, Junhui and Pan, Lihang and others},
  journal={arXiv preprint arXiv:2507.01006},
  year={2025}
}

@article{singh2025openai,
  title={Openai gpt-5 system card},
  author={Singh, Aaditya and Fry, Adam and Perelman, Adam and Tart, Adam and Ganesh, Adi and El-Kishky, Ahmed and McLaughlin, Aidan and Low, Aiden and Ostrow, AJ and Ananthram, Akhila and others},
  journal={arXiv preprint arXiv:2601.03267},
  year={2025}
}

@inproceedings{devlin2019bert,
  title={Bert: Pre-training of deep bidirectional transformers for language understanding},
  author={Devlin, Jacob and Chang, Ming-Wei and Lee, Kenton and Toutanova, Kristina},
  booktitle={Proceedings of the 2019 conference of the North American chapter of the association for computational linguistics: human language technologies, volume 1 (long and short papers)},
  pages={4171--4186},
  year={2019}
}

@article{dosovitskiy2020image,
  title={An image is worth 16x16 words: Transformers for image recognition at scale},
  author={Dosovitskiy, Alexey and Beyer, Lucas and Kolesnikov, Alexander and Weissenborn, Dirk and Zhai, Xiaohua and Unterthiner, Thomas and Dehghani, Mostafa and Minderer, Matthias and Heigold, Georg and Gelly, Sylvain and others},
  journal={arXiv preprint arXiv:2010.11929},
  year={2020}
}

@article{ikeda2015data,
  title={Data resource profile: the Japan National Health and nutrition survey (NHNS)},
  author={Ikeda, Nayu and Takimoto, Hidemi and Imai, Shino and Miyachi, Motohiko and Nishi, Nobuo},
  journal={International Journal of Epidemiology},
  volume={44},
  number={6},
  pages={1842--1849},
  year={2015},
  publisher={Oxford University Press}
}

@article{heuer2015food,
  title={Food consumption of adults in Germany: results of the German National Nutrition Survey II based on diet history interviews},
  author={Heuer, Thorsten and Krems, Carolin and Moon, Kilson and Brombach, Christine and Hoffmann, Ingrid},
  journal={British journal of nutrition},
  volume={113},
  number={10},
  pages={1603--1614},
  year={2015},
  publisher={Cambridge University Press}
}

@article{finglas2014assessing,
  title={Assessing and improving the quality of food composition databases for nutrition and health applications in Europe: the contribution of EuroFIR},
  author={Finglas, Paul M and Berry, Rachel and Astley, Si{\^a}n},
  journal={Advances in Nutrition},
  volume={5},
  number={5},
  pages={608S--614S},
  year={2014},
  publisher={Oxford University Press}
}

@book{mccance2014mccance,
  title={McCance and Widdowson's The Composition of Foods},
  author={McCance, R.A. and Widdowson, E.M. and Institute of Food Research (Great Britain) and Public Health England and Royal Society of Chemistry (Great Britain)},
  isbn={9781849736367},
  series={McCance and Widdowson's},
  year={2014},
  publisher={Royal Society of Chemistry}
}

@article{venables2022data,
  title={Data resource profile: united Kingdom National diet and nutrition survey rolling programme (2008--19)},
  author={Venables, Michelle C and Roberts, Caireen and Nicholson, Sonja and Bates, Beverley and Jones, Kerry S and Ashford, Robert and Hill, Suzanne and Farooq, Anila and Koulman, Albert and Wareham, Nicholas J and others},
  journal={International Journal of Epidemiology},
  volume={51},
  number={4},
  pages={e143--e155},
  year={2022},
  publisher={Oxford University Press}
}

@article{hu2022lora,
  title={Lora: Low-rank adaptation of large language models.},
  author={Hu, Edward J and Shen, Yelong and Wallis, Phillip and Allen-Zhu, Zeyuan and Li, Yuanzhi and Wang, Shean and Wang, Liang and Chen, Weizhu and others},
  journal={Iclr},
  volume={1},
  number={2},
  pages={3},
  year={2022}
}

@misc{google2025gemini3,
  title        = {Gemini 3},
  author       = {{Google DeepMind}},
  year         = {2025},
  month        = nov,
  howpublished = {\url{https://blog.google/products/gemini/gemini-3/}},
  note         = {Large language model. Released November 18, 2025}
}

@misc{anthropic2025sonnet45,
  title        = {Claude Sonnet 4.5},
  author       = {{Anthropic}},
  year         = {2025},
  month        = sep,
  howpublished = {\url{https://www.anthropic.com/news/claude-sonnet-4-5}},
  note         = {Large language model. Released September 29, 2025}
}

@article{cai2025z,
  title={Z-image: An efficient image generation foundation model with single-stream diffusion transformer},
  author={Cai, Huanqia and Cao, Sihan and Du, Ruoyi and Gao, Peng and Hoi, Steven and Hou, Zhaohui and Huang, Shijie and Jiang, Dengyang and Jin, Xin and Li, Liangchen and others},
  journal={arXiv preprint arXiv:2511.22699},
  year={2025}
}

@misc{labs2025flux1kontextflowmatching,
      title={FLUX.1 Kontext: Flow Matching for In-Context Image Generation and Editing in Latent Space},
      author={Black Forest Labs and Stephen Batifol and Andreas Blattmann and Frederic Boesel and Saksham Consul and Cyril Diagne and Tim Dockhorn and Jack English and Zion English and Patrick Esser and Sumith Kulal and Kyle Lacey and Yam Levi and Cheng Li and Dominik Lorenz and Jonas Müller and Dustin Podell and Robin Rombach and Harry Saini and Axel Sauer and Luke Smith},
      year={2025},
      eprint={2506.15742},
      archivePrefix={arXiv},
      primaryClass={cs.GR},
      url={https://arxiv.org/abs/2506.15742},
}

@inproceedings{zheng2024llamafactory,
  title={Llamafactory: Unified efficient fine-tuning of 100+ language models},
  author={Zheng, Yaowei and Zhang, Richong and Zhang, Junhao and Ye, Yanhan and Luo, Zheyan},
  booktitle={Proceedings of the 62nd annual meeting of the association for computational linguistics (volume 3: system demonstrations)},
  pages={400--410},
  year={2024}
}

\end{document}